\documentclass[sigconf,natbib=true]{acmart}
\usepackage{multirow}
\usepackage{multicol}
\usepackage{amsthm}
\usepackage{algorithmic}
\usepackage{bm} 
 \usepackage[normalem]{ulem}
\usepackage{booktabs}
\usepackage{subfigure}
\usepackage{enumitem}
\usepackage{amsmath,dsfont,bbm}
\usepackage{tcolorbox} 
\usepackage{fancybox}
\usepackage{fancyvrb}
\usepackage{xcolor}
\usepackage{dcolumn}
\definecolor{myboxcolor1}{RGB}{255,165,0}
\definecolor{myboxcolor2}{RGB}{255,200,100}
\usepackage[ruled,vlined]{algorithm2e}
\usepackage{tikz}
\usepackage{caption}
\usepackage{soul}

\AtBeginDocument{%
  \providecommand\BibTeX{{%
    \normalfont B\kern-0.5em{\scshape i\kern-0.25em b}\kern-0.8em\TeX}}}

\acmConference[arxiv]{}{}{}

\begin{document}
\title{Re2LLM: Reflective Reinforcement Large Language Model for Session-based Recommendation} 
\author{Ziyan Wang}
\email{wang1753@e.ntu.edu.sg}
\affiliation{%
  \institution{Nanyang Technological University}
  \country{Singapore}
}
\author{Yingpeng Du}
\email{dyp1993@pku.edu.cn}
\affiliation{%
  \institution{Nanyang Technological University}
  \country{Singapore}
}
\author{Zhu Sun}
\email{sunzhuntu@gmail.com}
\affiliation{%
  \institution{Agency for Science, Technology and Research}
  \country{Singapore}
}
\author{Haoyan Chua}
\email{haoyan001@e.ntu.edu.sg}
\affiliation{%
  \institution{Nanyang Technological University}
  \country{Singapore}
}
\author{Kaidong Feng}
\email{kaidong3762@gmail.com}
\affiliation{%
  \institution{Yanshan University}
  \country{China}
}
\author{Wenya Wang}
\email{wangwy@ntu.edu.sg}
\affiliation{%
  \institution{Nanyang Technological University}
  \country{Singapore}
}
\author{Jie Zhang}
\email{zhangj@ntu.edu.sg}
\affiliation{%
  \institution{Nanyang Technological University}
  \country{Singapore}
}

\begin{abstract}

 Large Language Models (LLMs) are emerging as promising approaches to enhance session-based recommendation (SBR), where both prompt-based and fine-tuning-based methods have been widely investigated to align LLMs with SBR.  
 However, the former methods struggle with optimal prompts to elicit the correct reasoning of LLMs due to the lack of task-specific feedback, leading to unsatisfactory recommendations. 
 Although the latter methods attempt to fine-tune LLMs with domain-specific knowledge, they face limitations such as high computational costs and reliance on open-source backbones. 
 To address such issues, we propose a \underline{Re}flective \underline{Re}inforcement \underline{L}arge \underline{L}anguage \underline{M}odel (Re2LLM) for SBR,  
guiding LLMs to focus on specialized knowledge essential for more accurate recommendations effectively and efficiently.   
 In particular, we first design the Reflective Exploration Module to effectively extract knowledge that is readily understandable and digestible by LLMs.  
 To be specific, we direct LLMs to examine recommendation errors through self-reflection and construct a knowledge base (KB) comprising hints capable of rectifying these errors.   
 To efficiently elicit the correct reasoning of LLMs, we further devise the Reinforcement Utilization Module to train a lightweight retrieval agent.   
 It learns to select hints from the constructed KB based on the task-specific feedback, where the hints can serve as guidance to help correct LLMs reasoning for better recommendations.  Extensive experiments on multiple real-world datasets demonstrate that our method consistently outperforms state-of-the-art methods.

\end{abstract}

\maketitle
\section{Introduction} 
Session-based recommendation (SBR) \cite{ intro1, intro2, autogsr, intro3, loam, attenmixer} plays a crucial role in real-world applications. It aims to capture users' dynamic preferences and predicts the next item that users may prefer based on previous interactions within a session. 
However, the user-item interactions are {often scarce}, and users' profiles are inaccessible in anonymous sessions, hindering the accurate recommendation result due to {data sparsity} and cold-start issues \cite{coldstart, sun2019research}. 

\begin{figure}[t] \centering
 \includegraphics[width=7cm]{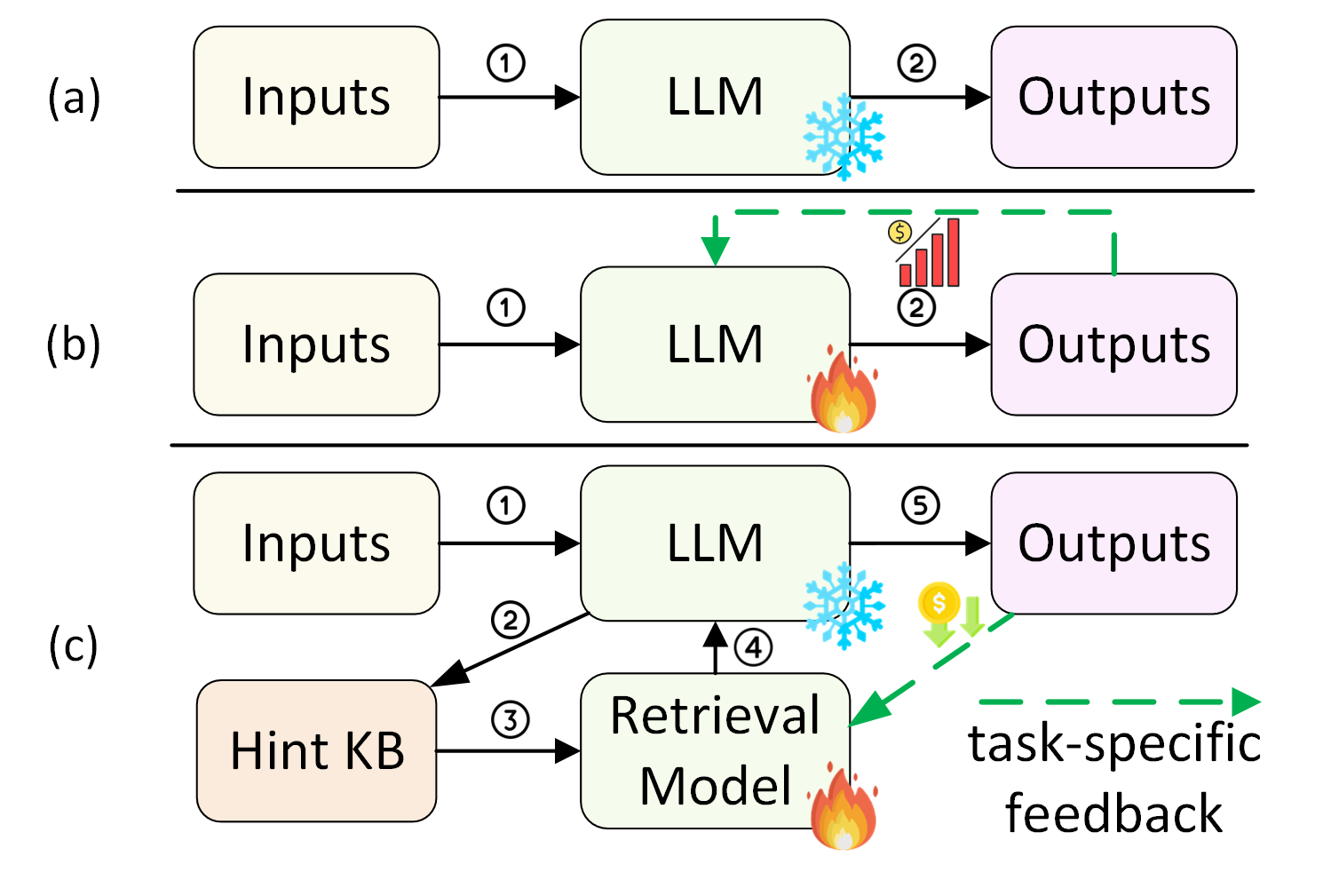} 
 \vspace{-10pt}
\caption{Our method (c) obtains effective task-specific feedback compared with prompt-based (a) and fine-tuning-based (b) methods.}\label{fig_intro}
\vspace{-18pt}
\end{figure}

Large Language Models (LLMs) have emerged to show potential in addressing these issues with their extensive knowledge and sophisticated reasoning capabilities. 
Recently, numerous methods have integrated LLMs into recommender systems (RSs), primarily through two ways: prompt-based methods and fine-tuning-based methods. 
The former methods exploit in-context learning and prompt optimization \cite{llmrank, chatrec, nir, sun2023large} to engage LLMs (e.g., ChatGPT\footnote{https://chat.openai.com/}) as recommenders without training as shown in Fig.\ref{fig_intro} (a). However, the crafted prompts not only require extensive expert knowledge and human labor but also may not align well with LLMs' understanding of SBR tasks, making LLMs vulnerable to hallucinations without effective supervision.
The latter methods focus on fine-tuning LLMs \cite{p5, palr, tallrec, LMasRS} with domain-specific knowledge, such as user-item interactions, in a supervised manner as shown in Fig.\ref{fig_intro} (b).
However, such methods often suffer from high computational costs, catastrophic forgetting, and reliance on open-source backbones.   
These drawbacks restrict the practical application of existing LLM-based methods for SBR.  

In this paper, we propose to direct LLMs to effectively and efficiently leverage specialized knowledge for recommendation without costly and inconvenient fine-tuning.
However, there remain two primary challenges to this goal:
(1) How can we effectively extract and craft specialized knowledge embedded within extensive user-item interactions to better align with LLM comprehension?
(2) How can we enable LLMs to utilize the specialized knowledge efficiently for better recommendations, without relying on resource-intensive supervised fine-tuning?

To overcome these challenges, we propose a \underline{Re}flective \underline{Re}inforce-ment \underline{L}arge \underline{L}anguage \underline{M}odel (Re2LLM) for SBR, aiming to capture and utilize specialized knowledge embedded in the extensive user-item interaction data by LLMs for more accurate recommendation. The proposed method consists of two main components: the Reflective Exploration Module and the Reinforcement Utilization Module.

\begin{itemize}[leftmargin=*]
    \item \textbf{Reflective Exploration Module} leverages the self-reflection of LLMs \cite{selfrefine, correctionsurvey} to extract specialized knowledge that is comprehensible and digestible by LLMs for SBR. Specifically, we employ LLMs to identify common errors in their responses, and then generate corresponding specialized knowledge (i.e., hints) to rectify these errors through LLMs' self-reflections. Hence, the specialized knowledge can align seamlessly with LLM comprehension as it is summarized by the LLM itself. Besides, we construct a hint knowledge base that serves as a pool to maintain specialized knowledge (hints), using an automated filtering strategy to ensure the effectiveness and diversity of retained hints. 
    \item \textbf{Reinforcement Utilization Module} employs a lightweight retrieval agent to select relevant specialized knowledge, thus eliciting the correct reasoning of LLMs without costly fine-tuning. Specifically, the retrieval agent is trained to select accurate hints from the knowledge base, which can prevent LLMs from potential errors in their inference process.  To overcome the absence of explicit labels about the effects of retrievals, we employ deep reinforcement learning (DRL) to simulate the real-world RSs for the agent. In detail, we first design an agent that selects hints (i.e., action) from the hint knowledge base by considering the session's contextual information (i.e., observation state). Then, we measure the improvement (i.e., reward) of recommendation results owing to the selected hint, and use them (rewards, observation states, and actions) as the task-specific feedback to update the agent via Proximal Policy Optimization (PPO) \cite{ppo} strategy.
\end{itemize}

In summary, as illustrated in Fig.\ref{fig_intro} (c), we develop a new learning paradigm to direct LLMs {to effectively explore specialized knowledge and efficiently utilize it for more accurate SBR.
Although the LLM backbone remains frozen, it can be guided by a lightweight retrieval that consumes task-specific feedback without costly fine-tuning. 
Therefore, our method combines the strengths of the large-scale LLM's effective reasoning capabilities and the efficiency of the lightweight retrieval model training. The key contributions of this paper are three-fold: 
\begin{itemize}[leftmargin=*]
\item 
We propose a new learning paradigm beyond in-context learning and fine-tuning of LLMs, {which seamlessly bridges between general LLMs and specific recommendation tasks.  It alleviates the issues of unaligned knowledge and costly supervised fine-tuning for LLMs, contributing to better SBR predictions.}
\item 
Our Re2LLM benefits from LLMs' self-reflection capabilities (i.e., Reflective Exploration module) as well as the flexibility of the lightweight retrieval agent (Reinforcement Utilization Module). {This enables us to effectively extract specialized knowledge understandable by LLMs, which can be then efficiently utilized for better LLM inference by learning from task-specific feedback.}
\item  Our extensive experiments demonstrate that Re2LLM outperforms state-of-the-art methods, including deep learning-based models and LLM-based models, in both few-shot and full-data settings across two real-world datasets.
\end{itemize}

\section{Literature Review}
\subsection{Session-based Recommendation}
SBR methods learn to model users' preferences and make recommendations based on short and dynamic sessions. 
The early classic work FPMC \cite{fpmc} combines Matrix Factorization \cite{koren2009matrix} and Markov Chain to capture sequential patterns and long-term user preference. With the development of deep learning technologies, various advanced techniques have been widely applied in SBR. 
\citet{DBLP:journals/corr/HidasiKBT15} first propose to use the Recurrent Neural Network (RNN) for SBR due to its strength in modeling sequential session interactions. 
Several methods propose variants by data augmentation \cite{augment}, novel ranking loss function \cite{lossfuc}, and mixture-channel purpose routing networks (MCPRN) \cite{mcprn}. 
Furthermore, the attention mechanism \cite{attentionalluneed} is introduced to capture more informative item representations for users' dynamic preferences for SBR, such as NARM \cite{narm} and STAMP \cite{stamp}. 
Recently, graph-based methods \cite{srgnn, gnn2, star, tagnn, hide} leverage the Graph Neural Network (GNN) to better learn high-order transitions within the sessions from the graph structure for SBR. 
For example, GCEGNN \cite{gcegnn} builds a global-level graph along with the item-level graph to exploit item transitions over all sessions for enhancement. 
Besides adopting different networks in SBR, many studies explore auxiliary information (e.g., attribute, description text) \cite{textsequence, decouplesideinfo, pricerec, knowledge-enhance} for better session profiling.
For example, MMSBR \cite{mmsbr} leverages both descriptive and numeric item attributes to characterize user intents. 
{Nevertheless, these methods may still suffer from inadequate user-item interactions and limited auxiliary information, hindering the accuracy of recommendation results.}

\subsection{Large Language Model for Recommendation} 
LLMs have emerged to show potential in addressing the aforementioned issues with their extensive knowledge and sophisticated reasoning capability, gaining their popularity for recommendation tasks.
Among LLM-based recommendation methods, most of the existing works attempt to utilize the LLM in two key strategies: prompt-based methods and fine-tuning (FT)-based methods. 
Prompt-based methods \cite{chatrec, llmfair, once, du2023enhancing, llmconversation, chatgptrs} retrieve information from recommendation data (e.g., historical user-item interactions) for direct outputs through prompt enhancement. 
LLMRank \cite{llmrank} and NIR \cite{nir} are two representatives that utilize prompt templates to extract dynamic preferences in anonymous sessions for SBR.
To enrich LLMs with task-specific supervision, FT-based methods adopt either fully \cite{p5, palr, llmunderstand, Li2023MultiModalityIA} or parameter-efficient (PEFT) approaches \cite{tallrec, genrec, llamarec,du2024large}, such as LoRA \cite{lora}, to fine-tune pre-trained LLMs for recommendation tasks. 
 For example, with PEFT, TALLRec \cite{tallrec} bridges the gap between LLMs and the recommendation tasks. PALR \cite{palr} fully fine-tunes LLaMA-7b \cite{llama} based on historical interactions to improve sequential recommendation performance.
 However, their effectiveness is constrained by the substantial computational demands, reliance on the availability of open-source LLM backbones, and inferior capabilities compared with larger-scale language models such as ChatGPT. 
{To address these challenges, we employ a lightweight retrieval agent to efficiently select relevant specialized knowledge summarized by LLMs, to elicit the correct reasoning of LLMs without costly fine-tuning.}

\subsection{Self-reflection in Large Language Models}
Recent advances in prompting strategies have effectively enhanced LLMs’ ability to handle complex tasks. 
Chain-of-Thought \cite{cot} and Tree-of-Thought \cite{tot} strategies allow LLMs to reason through a specified path. 
However, LLMs are still prone to hallucination and incorrect reasoning \cite{correctionsurvey}. 
Consequently, much effort is devoted to exploring LLMs' self-reflection \cite{promptopt,selfrefine,retroformer,reflexion} capabilities, where they iteratively refine outputs based on self-generated reviews or feedback without additional training. 
For example, \citet{selfrefine} apply this feedback-and-refine framework to improve LLMs' performance across NLP tasks. 
\citet{promptopt} improve prompts by having an LLM critique on generations and then refine them based on its feedback. 
\citet{retroformer} train a retrospective language model as a prompt generator, using the Proximal Policy Optimization (PPO) algorithm \cite{ppo} based on environment-specific rewards for optimization. 
However, the potential of LLMs' self-reflection in recommendation remains underexplored. 
To the best of our knowledge, only DRDT~\cite{drdt} is related to reflection for recommendation, which triggers LLMs to conduct the iterative reflection on a specific session until hitting the ground truth item. However, it is essentially a case-by-case reflection strategy, which hinders summarising the specialized knowledge from global sessions. To this end, we propose to explore different sessions to maintain a hint knowledge base, whose knowledge (i.e., hints) can be utilized to enhance LLMs reasoning for all sessions.

\section{Methodology}
\begin{figure*} \centering
 \includegraphics[width=0.75\textwidth]{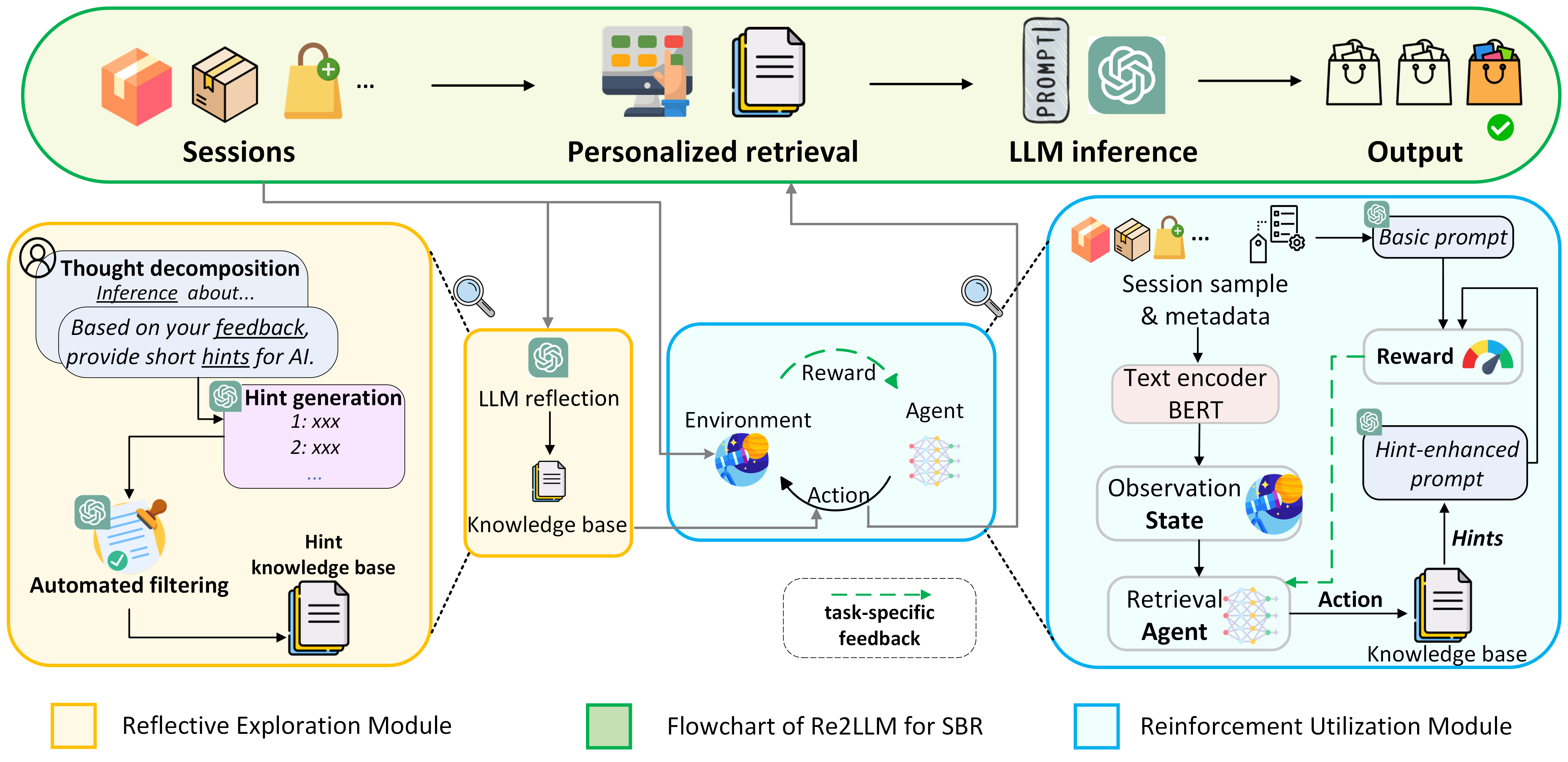}
 \vspace{-12pt}
\caption{The overall architecture of our proposed Re2LLM approach. The overall flowchart (green) contains a Reflective Exploration Module (yellow) for the generation of session-aware hints as specialized knowledge and a Reinforcement Utilization Module (blue) for the learning to retrieve obtained specialized knowledge.} \label{fig:main}
\vspace{-6pt}
\end{figure*}
In this section, we present our proposed Re2LLM approach, whose overall architecture is illustrated in Fig. \ref{fig:main}. Re2LLM comprises two key components: the Reflective Exploration Module and the Reinforcement Utilization Module. 
The Reflective Exploration Module facilitates the self-reflection of LLMs in identifying and {summarizing their inference errors for SBR, thereby generating hints (known as specialized knowledge) to avoid these potential errors}. These hints, further maintained by the automated filtering strategy to ensure their effectiveness and diversity, are stored in a knowledge base.
{The Reinforcement Utilization Module employs DRL to train a lightweight retrieval agent based on task-specific feedback without costly fine-tuning. The agent learns to select relevant hints to guide LLMs to mitigate potential errors in future inference.}
Finally, the inference is conducted by the LLM to deliver recommendations enhanced by these hints. 
For ease of presentation, we first introduce the notations used in this paper and provide problem formulation.

\medskip\noindent\textbf{Notations and Problem Formulation}.
Let $\mathcal{V} = \{v_1, v_2, ..., v_{N}\}$ denote the item set with $N$ items. 
Each session $s$ with $l$ interacted items is denoted as $s = \{v_1^s, v_2^s, ..., v_l^s\}$, where $v_i^s \in \mathcal{V}$. 
In addition, we are supposed to have side information on items, such as titles, genres, and actors of movies. 
Compared to traditional recommendation tasks, users' historical behaviors (i.e., interactions with items) on other sessions and their profiles are inaccessible in SBR due to the anonymous nature of sessions. 
The task of SBR is to predict the next item $v_{l+1}^s$ that the user is likely to interact with based on session history $s$.
For clarity, we denote the prompts designed in our methodology as $P_j$ where $j$ is the index. 
The notation $LLM(P_j)$ indicates the output of the LLM backbone given prompt $P_j$.

\subsection{Reflective Exploration Module}
Intuitively, accurate inference of LLMs requires domain-specific knowledge for recommendation tasks. However, the main challenge lies in that the domain-specific knowledge is usually embedded in the massive user-item interaction records, which may not well align with LLMs' comprehension.  
Alternatively stated, it is essential to extract the specialized knowledge that is intelligible for LLMs regarding recommendation tasks. 
Therefore, we propose to leverage LLMs' strong self-reflection capability for the extraction of specialized knowledge (i.e., hints) as LLMs can understand what they have summarized by themselves more easily (Section \ref{reflection}).
Additionally, we introduce an automated filtering strategy (Section \ref{filtering}) that maintains effective and diverse hints to construct a knowledge base for further utilization.

\subsubsection{Multi-Round Self-Reflection Generation} \label{reflection}
{To activate the self-reflection ability of LLMs, we propose to identify errors in LLMs' inference by comparing their predictions against the ground truth}. 
Specifically, we first instruct the LLM to generate a ranking list $O(s)$ from a candidate set $\mathcal{C}$ containing $\vert \mathcal{C}\vert$ items based on current session $s$:
\begin{equation}\label{eq:basic}
O(s) =  LLM(P_1(s, \mathcal{C})), O \subseteq \mathcal{C}, 
\end{equation}
where $P_1$ is the \textit{basic prompt} {consisting of titles of items that the user interacted with in the session $s$, candidate set $\mathcal{C}$, and task instruction. The details are shown in Prompt 1.}

\noindent{\shadowbox {\color{blue!30} { \small{\color{black}{\textbf{Prompt 1 (basic prompt)}}}}}

\noindent\shadowbox{ \label{basicprompt}
  \colorbox{yellow!10}{%
    \begin{minipage}{0.9\linewidth}
    \vspace{-3pt}
\small{\textit{I watched/purchased the following movies/items in order: \{\color{blue}$s$\color{black}\}. Based on these interactions, recommend a movie/item for me to watch/purchase next from a candidate set: \{\color{blue}$\mathcal{C}$\color{black}\}. Please recommend from the candidate set. List the top 10 recommendations in numbered bullet points.}}
\vspace{-3pt}
    \end{minipage}}}

    Then, we identify the incorrectly predicted sessions where the ranking list fails to hit the ground truth item at time step $l+1$ (i.e., $v_{l+1}^s\notin O(s)\O$). Afterward, we prompt the LLM to conduct self-reflection, that is, analyze and summarize the prediction errors, to generate potential hints in natural language to rectify such errors.
These hints thus can be considered as specialized knowledge for better recommendations.
Inspired by Chain-of-Thought prompting \cite{cot}, our approach focuses on analyzing and alleviating these errors in a step-by-step and progressive reasoning chain, producing hints that are tailored to be compatible with LLM comprehension capabilities, as introduced below.

\noindent{\shadowbox {\color{myboxcolor2} { \small{\color{black}{\textbf{Prompt 2}}}}}}

\noindent\shadowbox{ 
  \colorbox{yellow!10}{
    \begin{minipage}{0.9\linewidth}
    \vspace{-3pt}
    \small{
\textit{Question: \{\color{blue}$P_1$\color{black}\} (basic prompt). \newline ChatGPT: \{ \color{blue}$O(s)$\color{black}\} (top-10 results: 1. Casino Royale  2. Batman  3...). \newline
Now, know that none of these answers is the target. Infer about possible mistakes in the ChatGPT's predictions.
\newline 
\textbf{Example Output}:\newline
Here are potential mismatches or mistakes in the incorrect recommendations:\newline
\underline{Casino Royale}: While this is an action film, it leans more towards the spy genre, which isn't strongly represented in the watched list.\newline
\underline{Batman}: This recommendation assumes ...}}
\vspace{-3pt}
    \end{minipage}
  }}

\noindent{\shadowbox {\color{myboxcolor2} { \small{\color{black}{\textbf{Prompt 3}}}}}}

\noindent\shadowbox{ 
  \colorbox{yellow!10}{
    \begin{minipage}{0.9\linewidth}
    \vspace{-3pt}
\small{\textit{The correct answer is \{ \color{blue}{$v_{l+1}^s$}\color{black}\} (target). \newline Analyze why ChatGPT missed the target item from a wide variety of aspects. Explain the causes for the oversights based on earlier analysis. \newline 
\textbf{Example Output}:\newline
Given \color{blue} $v_{l+1}^s$ \color{black} is the correct item, we can infer several potential causes for the mismatch in the original recommendations:\newline
1. \textbf{Missing Subtle Preferences}: Your list contains a mixture of different genres, but there is a subtle preference for comedies. \newline 2. \textbf{Popularity Bias}: ...
}}
\vspace{-3pt}
    \end{minipage}%
  }
}

\noindent{\shadowbox {\color{myboxcolor2} {\small{ \color{black}{\textbf{Prompt 4}}}}}}

\noindent\shadowbox{ 
  \colorbox{yellow!10}{
    \begin{minipage}{0.9\linewidth}
    \vspace{-3pt}
\small{\textit{Provide short hints for AI to try again according to each point of the previous step, without information leakage of the target item.}}
\vspace{-3pt}
    \end{minipage}%
  }
}

Particularly, LLMs can yield incorrect results in SBR due to various causes, which underscores the importance of having {diverse and accurate hints to rectify the errors. 
To broaden the exploration of the diverse causes for the errors, we first only present the LLM with incorrectly predicted sessions and make it conduct self-reflection to identify possible causes for the errors without the ground truth item, as shown in Prompt 2. This allows the LLM to explore more diverse causes that may lead to errors.
Then, to pursue more accurate causes for the errors, we further stimulate the LLM's analytical skills by revealing the ground truth item.
Specifically, given the ground truth item, we ask the LLM to review previous diverse causes to select more relative ones, as shown in Prompt 3, thus identifying accurate causes of mistakes.
Finally, for effective utilization in the next stage, we ask the LLM to summarize obtained causes as hints (known as specialized knowledge) in Prompt 4, which are more understandable for LLMs and easily improve the recommendation outcome.}
By following the provided reasoning chain (i.e., Prompts 2-4), the LLM can identify frequent and prominent errors in SBR inference, and then produce qualified hints to address potential errors effectively.

\subsubsection{Automated Filtering} \label{filtering}
Recognizing that no single hint can rectify all the errors made by LLMs, we aim to build a hint knowledge base to store the most effective hints for further utilization, denoted as $\mathcal{H}$. 
To enhance the quality of the hints in $\mathcal{H}$, we develop an automated filtering strategy to maintain the hint knowledge base with two key properties: effectiveness and non-redundancy.

\noindent{\shadowbox {\color{blue!30} { \small{\color{black}{\textbf{Prompt 5 (hint-enhanced prompt)}}}}}

\noindent\shadowbox{ \label{basicprompt}
  \colorbox{yellow!10}{%
    \begin{minipage}{0.9\linewidth}
    \vspace{-3pt}
\small{\textit{I watched/purchased the following movies/items in order: \{\color{blue}$s$\color{black}\}. Based on these interactions, recommend a movie/item for me to watch/purchase next from a candidate set: \{\color{blue}$\mathcal{C}$\color{black}\}. Please recommend from the candidate set. List the top 10 recommendations in numbered bullet points. \textbf{Hint: \{\color{blue}$h$\color{black}\}} (retrieved hint).}} \vspace{-3pt}
    \end{minipage}%
  }
}

{To ensure the effectiveness, we only add a new hint to $\mathcal{H}$ if it can lead to performance improvement.
Specifically, we first construct a \textit{hint-enhanced prompt} $P^*$ (Prompt 5) to trigger the LLM with the hint-enhanced prompt for recommendation inference:
\begin{equation}
    O^*(s) =  LLM(P^*(s, \mathcal{C}, h)),
\end{equation}
where $O^*(s)$ is the recommendation list for session $s$ based on the hint-enhanced prompt $P^*$. 
{Then, we compare $O^*(s)$ with $O(s)$ obtained by Eq. \ref{eq:basic} using the basic prompt without hint enhancement. The hint $h$ can be recognized as effective if it leads to a better prediction by the LLM, i.e., $v_{l+1}^s \in O^*(s) \And v_{l+1}^s \notin O(s)$.}

\noindent{\shadowbox {\color{myboxcolor2} { \small{\color{black}{\textbf{Prompt 6 }}}}}

\noindent\shadowbox{ 
  \colorbox{yellow!10}{
    \begin{minipage}{0.9\linewidth}
    \vspace{-3pt}
\small{\textit{`Does hint [\color{blue}$h'$\color{black}] convey a similar idea for these hints: [\color{blue}$h$\color{black}]? Return 1 if true, else return 0.'}}
\vspace{-3pt}
    \end{minipage}
  }
}

Meanwhile, some hints may be redundant generated by LLMs, as multiple sessions may suffer from similar causes. For example, `\textit{Consider the release years of movies in the user’s history for era preference}' and `\textit{Think about the production year of watched movies, recommend movies in that era from the candidate set}' share a similar semantic meaning. 
The redundancy of hints may result in a high cost of the maintenance and utilization of the hint knowledge base $\mathcal{H}$, e.g., larger size of the hint knowledge base and increased complexity in retrieving the targeted hint.
%
To reduce the redundancy in the hint knowledge base $\mathcal{H}$, we only incorporate new hints that are distinct from existing ones. 
Specifically, we employ LLMs to detect the semantic similarity between the candidate hint $h'$ and existing ones $\{h\in \mathcal{H}\}$, given by,
\begin{equation}
     \sum\nolimits_{h\in \mathcal{H}} LLM(P_6(h',h))== 0, \label{eq:topic}
 \end{equation}
where the detail of $P_6$ is shown in Prompt 6, instructing the LLM to return 0 if $h'$ shares a different idea with $h$ and else return 1. By doing this, only distinct candidate hint $h'$ will be incorporated into the knowledge base $\mathcal{H}$ if it satisfies Eq. \ref{eq:topic}.

Through the automated filtering process, we iteratively update the hint knowledge base until reaching capacity. 
It contains qualified hints to correct the errors across different session patterns and thus improve recommendation results. 
For a practical illustration, we randomly select five hints from the movie and the video game domains in the diagrams below. 
\definecolor{myboxcolor2}{RGB}{255,200,100}
\newtcolorbox{mybox3}[1]{
  colback=myboxcolor2!5!white,
  colframe=myboxcolor2!75!black,
  fonttitle=\bfseries,
  title=#1,
  halign title=center, 
  halign=flush left,   
  left=1mm,
  right=1mm
}

\begin{mybox3}{Hint examples - Movie}
\vspace{-5pt}
\small{\textit{-\quad Consider the release years of movies in the record for era preference.\\
-\quad Pay attention to actors who appear in multiple movies from the watched list and consider other films featuring these actors.\\
-\quad Aim for a personalized recommendation list that offers variety and reflects a range of moods and themes.\\
-\quad Consider films that originated from other media, such as books, radio plays, or TV shows. \\
-\quad Do not rely solely on mainstream popularity; consider critically acclaimed films that may not be box office hits.}}
\vspace{-5pt}
\end{mybox3}
\vspace{-10pt}
\begin{mybox3}{Hint examples - Video Game}
\vspace{-5pt}
\small{\textit{-\quad Pay attention to the platforms prominently featured in the user's purchase history and include items that are relevant to those platforms.\\
-\quad Give preference to games released in the era aligning with purchased games.\\
-\quad Pay attention to the previous purchase history, which indicates an interest in gaming accessories and hardware enhancements.\\
-\quad Ensure the games are suitable for a wide range of ages and have a universal appeal. \\
-\quad Focus on items that specifically enhance or are used in conjunction with mobile devices.}}
\vspace{-5pt}
\end{mybox3}

\subsection{Reinforcement Utilization Module}
To guide LLMs to infer more accurate SBR predictions, we propose to utilize the constructed hint knowledge base to prevent errors in the inference process of LLMs. 

In practical scenarios, the absence of explicit labels on the hints' efficacy leads to a challenge in conducting supervision on hint selection, as computing the efficacy for all hints on each sample is costly and redundant. Fortunately, DRL with a replay buffer enables us to collect reward signals (i.e., hint efficacy) and update the network spontaneously to speed up.
To this end, we employ the Proximal Policy Optimization (PPO) \cite{ppo} algorithm to ensure stable and efficient training of our retrieval agent. 
Specifically, our agent is trained to select the most relevant hints based on session-related context information, thereby preventing LLMs from similar reasoning mistakes for recommendation. In addition, the proposed DRL framework allows for the balance between exploitation and exploration. It is also compatible with more complex extensions such as retrieving multiple hints and multi-round agent-based interactions.

\subsubsection{Markov Decision Process (MDP)}
To outline the basics of our DRL environment, we define MDP as the tuple $MDP = (\mathcal{Z}, \mathcal{A}, R, T)$, where $\mathcal{Z}$, $\mathcal{A}$, $R$ 
and $T$ denote the state space, action space, reward function, and transition, respectively. 

\medskip\noindent\textbf{State}. To model the session-related context information for our retrieval agent, we concatenate item titles and attributes into a single text string and convert it into embedding for semantic extraction. 
We use pre-trained BERT \cite{bert} as the text encoder due to its robust contextual language understanding capabilities. 
The state can be denoted as $\mathbf{z}= BERT(s)$, where $\mathbf{z} \in \mathcal{Z}$ is the $d$-dimensional output of the text encoder. 

\medskip\noindent\textbf{Action}. To select relevant hints from the constructed knowledge base for LLMs' inference, we define a discrete action space for the agent, represented as $\mathbf{a} \in\mathcal{A}$, which is a $(|\mathcal{H}|+1)$-dimensional vector. 
Here, $|\mathcal{H}|$ is the size of the knowledge base, and an action $\mathbf{a}$ corresponds to either choosing a hint or opting not to select any. 

\medskip\noindent\textbf{Reward}. To direct the agent in making accurate hint selections, we employ a comparative function $R$ to provide the reward signals for the agent's actions. This function evaluates the improvement in the LLM prediction accuracy by the hint-enhanced prompt (Prompt 5) versus the basic prompt (Prompt 1). 
For each episode, the reward value $r$ is denoted as $r= m(O^*(s)) - m(O(s))$, where $m$ is a recommendation evaluation metric (e.g., NDCG@10). 

\medskip\noindent\textbf{Transition}.  
{In each step, the agent can receive task-specific feedback by simulating the real-world RSs. The agent observes the state of the current session, takes action to select a hint for the LLM, and receives a reward from the environment.}
Subsequently, the environment transits to the next session's state, and then the transition is denoted as $T(\mathbf{z}) = \mathbf{z}' $. This setup can flexibly accommodate 
the scenario where 
multiple steps per session are involved for more sophisticated learning processes in our future studies.

\medskip\noindent\textbf{Replay buffer}. To facilitate efficiency in policy optimization, we maintain a replay buffer $D = (\mathbf{z}, \mathbf{a}, r, \mathbf{z}')$ to store the tuples of observation state, agent action, reward, and next observation state. 
With the records in the replay buffer, the retrieval agent can refine successful strategies and learn from erroneous trials.
This technique significantly accelerates the DRL training as the LLM backbone is relatively slow for inference and producing reward signals.

\subsubsection{PPO Training}
To model the actions of the retrieval agent, we implement a policy network parameterized by MLPs to define our agent’s policy $\pi_{\bm{{\theta}}}$, which maps the environmental space $\mathcal{Z}$ to the action space $\mathcal{A}$:
\begin{equation}
\mathbf{a} = softmax(\mathbf{W}\cdot \mathbf{z}), \label{eq:softmax}
\end{equation}
where $\mathbf{W}\in\mathbb{R}^{(|H|+1)\times d}$ is the learnable weight matrix, and $\mathbf{z} \in\mathbb{R}^{d}$ denotes the state of the current session. The action of the agent corresponds to the largest value among the softmax logits of $\mathbf{a}$. 

During the training process, the retrieval agent adopts the $\epsilon$-greedy ($\epsilon = 0.05$) strategy to explore the environment, with probability $\epsilon$ to take a random action while with probability ($1-\epsilon$) to exploit the learned policy $\pi_{\bm{\theta}}$. 
Our objective function for PPO training can be formulated to maximize the total reward, given by,
\begin{equation}
\mathcal{L}_{PPO} = \mathbb{E}\left[ R(\mathbf{z}, a) - \beta KL(\pi_{{\bm{\theta}_{old}}},\pi_{\bm{{\theta}}}) \right], \label{eq:ppo}
\end{equation}
where the first term is the reward measured by recommendation tasks, and the second term is the KL-divergence of policies with a coefficient $\beta$ for the regulation in policy updates.  
{In summary, we train a retrieval agent with task-specific feedback via DRL, which can select the appropriate action to improve LLM's performance through hint-enhanced prompts for better recommendations.}
\begin{algorithm}[t]
\small
\KwIn{Training data $\mathcal{S}_{tr}$ for SBR, hint knowledge base size $n_h$.}
\tcp{Reflective Exploration Module}
\While{$|\mathcal{H}| < n_h$}{
    Sample $s$ from $S_{tr}$\;
    \If{m(O($s$))=0    \tcp{If incorrectly predicted}
}{
        Obtain multiple hints by \textit{Prompt 2 - 4}\;
        \For{each obtained hint $h$}{
            Assess two aspects of quality by automated filtering\;
            \If{True}{
                Add $h$ to $\mathcal{H}$\;
            }
        }
    }
}
\Return $\mathcal{H}$ 

\vspace{-0.1in}

\begin{tikzpicture}
\vspace{-10pt}
    \draw[dotted] (0,0) -- (\linewidth,0);
\end{tikzpicture}

\vspace{-0.05in}

\KwIn{Training data $\mathcal{S}_{tr}$ for SBR, constructed $\mathcal{H}$, max episode $N$.}
Initialize policy network $\pi_{\bm{\theta}}$\;
\tcp{Reinforcement Utilization Module}
\For{$i = 1, 2, ..., N$}{
    Shuffle $\mathcal{S}_{tr}$\;
    \For{$s$ in $\mathcal{S}$}{
        Encode $\mathbf{z} \in \mathcal{Z}$ based on context information of $s$\;
        Sample action $a$ by Eq. \ref{eq:softmax} and $\epsilon$-greedy exploration\;
        Compute reward by $r = m(O^*(s)) - m(O(s))$\;
        Transit to the next state $\mathbf{z}'$\;
        Store the tuple $(\mathbf{z}, \mathbf{a}, r, \mathbf{z}')$ into the replay buffer\;
    }
    Sample from the replay buffer, update policy by Eq. \ref{eq:ppo}\;
}
\Return $\pi_{\bm{\theta}}$ 
\caption{Re2LLM}
\label{algo2}
\vspace{-1pt}
\end{algorithm}
\subsubsection{Retrieval-Enhanced LLMs for Recommendation}
With the constructed knowledge base and the trained retrieval agent, we achieve more accurate recommendations  by automatically
selecting appropriate hints for prompt enhancement during inference. 
The final recommendation output $\hat{O}$ can be denoted as follows:
\begin{equation}
    \hat{O} =  LLM(P^*(s, \mathcal{C}, \pi_{\bm{\theta}}(\mathbf{z}_s))), \label{inferr}
\end{equation}
where $\pi_{\bm{\theta}(\mathbf{z_s})}$ corresponds to the selected hint $h$ by the trained retrieval agent for session $s$ with its state $\mathbf{z}_s$. 
The overall algorithm of the proposed method Re2LLM is shown in Algorithm \ref{algo2}.

\section{Experimental Results}
In this section, we evaluate the effectiveness of the proposed method Re2LLM through comprehensive experiments and analysis\footnote{Our source code is available at  \url{https://anonymous.4open.science/r/Re2LLM-34DC/}.}.
We aim to answer the following research questions:
\begin{itemize}[leftmargin=*]
\item  \textbf{RQ1}: Whether the proposed method outperforms baseline methods, including the deep learning models and other LLM-based models, for SBR?
\item  \textbf{RQ2}: How can key components affect our proposed method? Specifically, how is the efficacy of the proposed Reflective Exploration Module and Reinforcement Utilization Module?
\item \textbf{RQ3}: How do key hyper-parameters impact the performance of our proposed method?
\end{itemize}
\begin{table}[t]
\caption{Statistics of datasets.} \centering
\label{Table_es}
\vspace{-0.16in}
\begin{tabular}{c|c|c}
\toprule
& \textbf{Movie} & \textbf{Game} \\
\midrule
\# Sessions & 468,389 & 387,906 \\
\# Items & 8,233 & 22,576 \\
Avg. length & 10.17 & 6.28 \\
\hline
Utilized Item & title, genre, actor, year, & title, category, tag, \\
Side Information & country, director & brand, description\\
\bottomrule
\end{tabular}
\vspace{-15pt}
\end{table}

\begin{table*}[htbp!]
\renewcommand{\arraystretch}{1}
\setlength\tabcolsep{6.5pt}
\centering
\fontsize{8.5}{8.5}\selectfont
\caption{
  Performance of all methods under two settings. The performance of our Re2LLM is in bold; the best performance achieved by baselines in full and few-shot settings are underlined with `\underline{ \; }' and `\uuline{ \; }', respectively. 
  The improvements (Imp.) of our model against the best-performed baselines in both settings are reported, where * and {$\dagger$} indicate statistically significant improvement by t-test (p < 0.05) for full* and few-shot$^\dagger$ settings, respectively.}
  \label{tab:mainresult}
  \vspace{-0.1in}
\begin{tabular}{c|c|cccc|cccc}
\toprule
\multirow{2}[3]{*}{\textbf{Setting}}& \multirow{2}[3]{*}{\textbf{Model}} & \multicolumn{4}{c|}{\textbf{Movie}} & \multicolumn{4}{c}{\textbf{Game}} \\ 
     \cmidrule{3-10}
     & & \multicolumn{1}{c}{HR@5} & \multicolumn{1}{c}{HR@10}& \multicolumn{1}{c}{NDCG@5} & \multicolumn{1}{c|}{NDCG@10} & \multicolumn{1}{c}{HR@5} & \multicolumn{1}{c}{HR@10}  & \multicolumn{1}{c}{NDCG@5} & \multicolumn{1}{c}{NDCG@10}\\ 
     \midrule
 \multirow{7}{*}{full dataset} & FPMC  & 0.1492 & 0.2885 & 0.0922 & 0.1749 & 0.2463 & 0.3617 & 0.1788 & 0.2110    \\
 & GRU4Rec & 0.2516 & 0.3861 & 0.1564 & 0.2103 & 0.2915 & 0.4107 & 0.2004 & 0.2392 \\
& NARM & 0.3867 & 0.5674 & 0.2605 & 0.3319 & 0.4743 & 0.5635 & 0.3732 & 0.4021 \\
 & SRGNN & 0.3879 & 0.5720 & 0.2676 & 0.3342 & 0.4864 & 0.5335 & 0.3786 & 0.4128 \\
 & GCEGNN &  \underline{0.3962} & 0.5895 & 0.2659 & 0.3281 & 0.4826 & 0.5523 & \underline{0.4002} & \underline{0.4332} \\
 & AttenMixer & 0.3880 & \underline{0.5727} & \underline{0.2693} & \underline{0.3472} & \underline{0.4885} & \underline{0.5838} & 0.3911 & 0.4226 \\
\cmidrule{2-10}
&\textbf{Imp.} & $4.14\%^*$  & 0.14\%  & $5.42\%^*$  & -3.39\%  & $15.95\%^*$  & $22.87\%^*$  & $4.82\%^*$  & $8.24\%^*$ \\
\midrule
\multirow{10}{*}{few-shot}& NARM & 0.2701 & 0.4412 & 0.1707 & 0.2298 & 0.1891 & 0.2765 & 0.1310 & 0.1586 \\
& NARM-attr & 0.3108 & 0.4835 & 0.1920 & 0.2481 & 0.1977 & 0.2915 & 0.1372 & 0.1672 \\
 & GCEGNN & 0.3227 & 0.4813 & 0.2080 & 0.2584 & 0.1905 & 0.2798 & 0.1362 & 0.1659 \\
  & GCEGNN-attr & 0.3254 & 0.4893 & 0.2136 & 0.2678 & 0.2019 & 0.3992 & 0.1408 & 0.1710 \\
 & AttenMixer & 0.2774 & 0.4668 & 0.1774 & 0.2377 & 0.1899 & 0.2925 & 0.1357 & 0.1581 \\
 & AttenMixer-attr & 0.3397 & 0.5056 & 0.2223 & \uuline{0.2795} & 0.2084 & 0.3121 & 0.1371 & 0.1727 \\
 & LLMRank & \uuline{0.3649} & \uuline{0.5110} & \uuline{0.2436} & 0.2784 & 0.4947 & \uuline{0.6636} & 0.3898 & \uuline{0.4531} \\
 & NIR & 0.3587 & 0.4901 & 0.2214 & 0.2648 & \uuline{0.5291} & 0.6434 & \uuline{0.4016} & 0.4346 \\
 \cmidrule{2-10}
 & \textbf{Re2LLM} & \textbf{0.4126} & \textbf{0.5735} & \textbf{0.2839} & \textbf{0.3358} & \textbf{0.5664} & \textbf{0.7173}& \textbf{0.4195} & \textbf{0.4689} \\
 \cmidrule{2-10}
 & \textbf{Imp.} & $13.07\%^\dagger$ & $12.23\%^\dagger$  & $16.54\%^\dagger$  & $20.14\%^\dagger$  & $7.05\%^\dagger$  & $8.09\%^\dagger$  & $4.46\%^\dagger$  & $3.49\%^\dagger$ \\
\bottomrule
\end{tabular}
\vspace{-10pt}
\end{table*}

\subsection{Experimental Setup}
\subsubsection{Dataset.}
In this paper, we evaluate the proposed Re2LLM and baseline methods on two real-world datasets, namely Hetrec2011-Movielens and Amazon Game:
\begin{itemize}[leftmargin=*]
\item Hetrec2011-Movielens\footnote{https://grouplens.org/datasets/hetrec-2011/} dataset contains user ratings of movies. It also contains the side information of movies, e.g., title, production year, and movie genre. 
\item Amazon Game\footnote{https://jmcauley.ucsd.edu/data/amazon/} dataset is the `Video Game' category of the Amazon Review Dataset, which is collected from the Amazon platform with users’ reviews on various types of games and peripherals. It also contains metadata of games, e.g., title, brand, and tag.
\end{itemize}

For each user, we take all the interactions within one day as a session by the recorded timestamps. 
We filter out sessions and items that have less than 3 records and treat records with ratings as positive implicit feedback. 
Following the prior study \cite{augment}, we adopt the data augmentation strategy to extend a session $s$ with length $\vert s\vert$ to $(\vert s\vert-1)$ sessions as training samples.
The statistics of the datasets are summarized in Table \ref{Table_es}.

\subsubsection{Baselines}
We compare our proposed method with eight state-of-the-art baseline methods\footnote{Our method does not require fine-tuning, and due to the high computational cost, fine-tuning-based methods are not in our scope. }, and the details are listed as follows. \textbf{FPMC} \cite{fpmc} combines matrix factorization with the first-order Markov chain.    
\textbf{GRU4Rec} \cite{DBLP:journals/corr/HidasiKBT15} uses Gated Recurrent Unit (GRU), a typical RNN layer, to model whole sessions.  
\textbf{NARM} \cite{narm} introduces the attention mechanism into RNNs to better model the sequential behavior. 
\textbf{SRGNN} \cite{srgnn} constructs session graphs and employs GNN layers to obtain embeddings. 
\textbf{GCEGNN} \cite{gcegnn} further includes global-level information as an additional graph to enhance the model performance. 
\textbf{AttenMixer} \cite{attenmixer} achieves multi-level reasoning over item transitions by the attention-based readout method.    
\textbf{LLMRank} \cite{llmrank} is the LLM-based method, which demonstrates the promising zero-shot inference ability of LLMs through recency-focused prompting and in-context learning. 
\textbf{NIR}~\cite{nir} is also LLM-based method that employs a multi-step prompt to capture user preferences first, then select representative movies, and finally perform the next-item prediction. 

{For a comprehensive comparison, we introduce two settings for the implementation of these baseline methods. }
In the \textit{full dataset setting}, we investigate how Re2LLM performs in comparison with baseline methods trained on the full augmented dataset using item IDs.
In the \textit{few-shot setting}, we examine whether our Re2LLM shows superior performance compared to the baselines trained with limited data\footnote{We select more representative and stronger baselines for the few-shot setting and incorporate item attribute information in the few-shot setting only for a fair comparison.}.
{The `-attr' suffix denotes the incorporation of item attributes for deep learning-based baselines. We first concatenate all attributes (title included) of the item and then encode them into a text-aware embedding by BERT. Finally, we further aggregate the ID embedding and text-aware embedding of the item as its underlying representation.} 
\renewcommand{\arraystretch}{1}

\subsubsection{Evaluation Strategy and Metrics}
For each dataset, we apply the split-by-ratio strategy~\cite{daisyrec} for sessions to obtain training, validation, and test sets with a ratio of 7:1:2. 
For the full dataset setting, we use the entire training set. 
For the few-shot setting, we sample 500 training samples from the entire training set. 
{In terms of evaluation, we consider the last item in each session as the target. Considering the cost and efficiency of LLMs, we randomly sample a subset of 2,000 sessions from the test set. For the efficiency of evaluation, we sampled 49 negative items for each positive item (i.e., target). 
}  
In addition, we adopt two commonly used metrics for model evaluation, including normalized discounted cumulative gain ($NDCG@n$) and hit rate ($HR@n$), with $n=\{5, 10\}$. 
The results show the average of five runs with different random seeds.

\subsubsection{Implementation Details} 
{For a fair comparison, all methods are optimized by the Adam optimizer with the same embedding dimension $768$ that aligns with the BERT encoder dimension\footnote{Empirically, using MLP to reduce dimension leads to a consistent performance drop.}. Following \cite{daisyrec2}, we employ the Optuna\footnote{https://optuna.org/} framework to efficiently optimize the hyper-parameter of all methods. We conduct 20 trials on the following searching spaces, i.e., \textit{learning rate}: \{$1e^{-2},1e^{-3},1e^{-4}$\}, \textit{weight decay}: \{$1e^{-4},1e^{-6}$ $,1e^{-8}$\}, and \textit{batch size}: \{$16,64,256$\}. For our method Re2LLM, we set candidate set size $\vert \mathcal{C}\vert$ to 50, knowledge base size $n_h$ to 20, and few-shot training size $n_e$ to 500. The impacts of essential hyper-parameters on Re2LLM can be found in Section \ref{sec_hyper}.  For LLM-based methods, we use `gpt-4' API as the backbone model for these methods.  
For the rest hyper-parameters in baseline methods, we follow the suggested settings in the original papers. The experiments are run on a single Tesla V100-PCIE-32GB GPU.}  On average, around 50 data samples are required to generate 20 hints. The average cost to analyze and conduct inference on each sample is around USD 0.015 during the hint generation, DRL training, and final testing.
\begin{table*}[t]
  \centering
  \fontsize{8}{8}\selectfont
  \caption{The results of ablation study across all datasets.}
  \label{tab:ablation}
  \vspace{-0.15in}
\begin{tabular}{c|cccc|cccc}
\toprule
 \multirow{2}[3]{*}{\textbf{Models}} & \multicolumn{4}{c|}{\textbf{Movie}} & \multicolumn{4}{c}{\textbf{Game}} \\ \cmidrule{2-9}
 & \multicolumn{1}{c}{HR@5} & \multicolumn{1}{c}{HR@10}& \multicolumn{1}{c}{NDCG@5} & \multicolumn{1}{c|}{NDCG@10} & \multicolumn{1}{c}{HR@5} & \multicolumn{1}{c}{HR@10}  & \multicolumn{1}{c}{NDCG@5} & \multicolumn{1}{c}{NDCG@10}\\ 
 \midrule
w/o-HE  & 0.3923          & 0.5458          & 0.2704          & 0.3198          & 0.5462& 0.6771
          & 0.4182& 0.4550      \\
w/o-AF          & 0.4013        & 0.5592         & 0.2774          & 0.3289          & 0.5652         & 0.6923      & 0.4146          & 0.4602          \\
w/o-DRL(RAN)        & 0.3542          & 0.4960          & 0.2446          & 0.2908          & 0.5072          & 0.6345          & 0.3722          & 0.4107          \\
w/o-DRL(ALL)    &0.3910&	0.5434&	0.2746	&0.3235&0.5638&0.6897&0.4164&0.4592\\
\textbf{Re2LLM} & \textbf{0.4126} & \textbf{0.5735} & \textbf{0.2839} & 
   \textbf{0.3358} & \textbf{0.5664} & \textbf{0.7173} & \textbf{0.4195} & \textbf{0.4689}\\
\bottomrule
\end{tabular}
\vspace{-12pt}
\end{table*}
\begin{figure*}[htbp!] \centering
 \includegraphics[width=14cm]{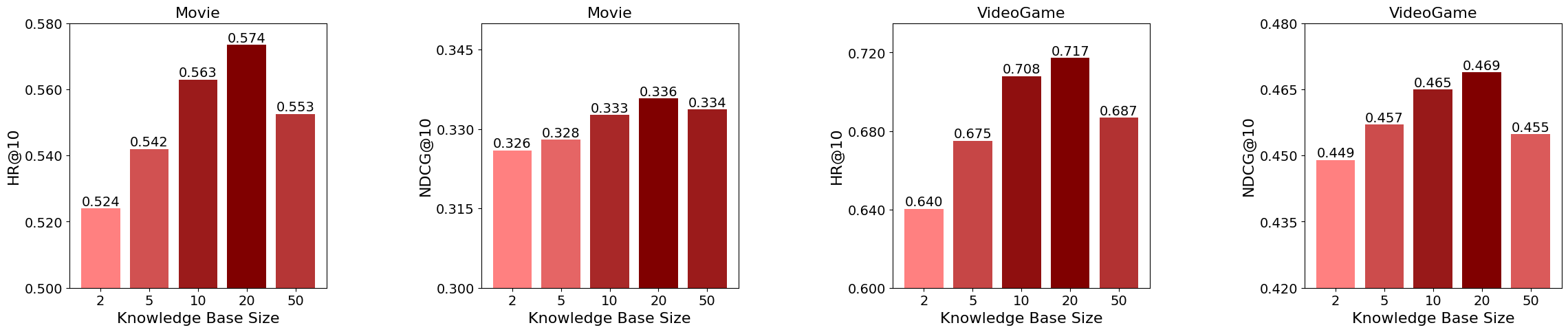}
 \vspace{-10pt}
\caption{Performance of the proposed method with varying knowledge base size. Darker color indicates higher value.}\label{fig_hint}
\vspace{-15pt}
\end{figure*}

\subsection{Overall Comparison (RQ1)}
Table \ref{tab:mainresult} shows the performance of our method Re2LLM and various baseline methods under two evaluation settings, i.e., full dataset setting and few-shot setting. 
{The experimental results lead to several key conclusions. 
\textbf{First}, our proposed method significantly outperforms baselines in few-shot settings, with average improvements of 15.49\% and 5.77\% across all metrics when compared to the top-performing baseline models on Movie and Game, respectively. This is mainly attributed to the effectiveness of Re2LLM, which not only extracts specialized and comprehensible knowledge by self-reflection but also effectively utilizes the knowledge based on a well-trained retrieval agent for better recommendations. \textbf{Second}, Re2LLM achieves comparable and even better performance than baseline methods trained on full datasets, which also verifies the effectiveness of our module design. Furthermore, the LLM-based methods (e.g., LLM-Ranker and NIR) also achieve comparable performance to baseline methods trained on full datasets, which indicates the promising way of employing LLMs as recommenders.}
\textbf{Third}, deep learning-based models show a significant performance drop in the few-shot setting compared with the full dataset setting, pointing to the data sparsity issue as a primary challenge. 
\textbf{Last}, the incorporation of item attributes slightly boosts the performance of baseline methods, showing the potential of using side information of items in addressing the sparsity issue in SBR.
\captionsetup[figure]{belowskip=-12pt, aboveskip=-3pt}

\subsection{Ablation Studies (RQ2)}
To verify the impact of each component, we compare our method Re2LLM with its variants:
\begin{itemize}[leftmargin=*]
\item w/o-HE: We only adopt the \textit{basic prompt} (Prompt 1) to trigger the LLM to generate recommendations instead of \underline{H}int-\underline{E}nhanced prompt as in Eq.~\ref{inferr}. 
\item w/o-AF: We remove the \underline{A}utomated \underline{F}iltering mechanism from the hint knowledge base construction, that is, incorporating all hints generated by LLMs into the hint knowledge base regardless of their quality.
\item w/o-DRL: We remove the \underline{D}eep \underline{R}einforcement \underline{L}earning with task-specific feedback for the retrieval agent. Instead, we use two straightforward strategies for hint retrieval, namely \underline{RAN}dom and \underline{ALL} retrieval strategies. Specifically, the variant w/o-DRL(RAN) randomly samples hints from the hint knowledge base for prompt enhancement, whereas the variant w/o-DRL(ALL) concatenates all hints in the knowledge base instead of the selective retrieval.
\end{itemize}

Table \ref{tab:ablation} shows the performance of all ablation studies. 
\textbf{First}, Re2LLM outperforms its variant w/o-HE, which indicates the necessity of activating LLM's reflection mechanism with hint enhancement in our Reflective Exploration Module. 
\textbf{Second}, Re2LLM outperforms variants w/o-DRL(RAN) and w/o-DRL(ALL), indicating the importance of our Reinforcement Utilization Module for the training of the retrieval agent with task-specific feedback signals. 
The variant w/o-DRL(RAN) has the worst performance by random sampling hints because inaccurate hints can mislead the LLM inference for recommendation, resulting in negative impacts. 
The variant w/o-DRL(ALL) shows limited improvement compared to the variant w/o-HE, revealing that simply feeding all hints into LLMs without session-aware selection is not optimal. 
\textbf{Third}, the variant w/o-AF achieves relatively better performance among all variants. Nevertheless, it still underperforms our {Re2LLM} due to the positive impact of the automated filtering mechanism on the hint knowledge base construction. 
In summary, the above results exhibit the efficacy of different modules in the proposed Re2LLM for more accurate SBR.

\subsection{Hyper-parameter Analysis (RQ3)}\label{sec_hyper}
We now examine the impact of several key hyper-parameters and designs on our proposed Re2LLM, including knowledge base size $n_h$, few-shot training size $n_e$, and reward design. 

\subsubsection{Knowledge Base Size}
We investigate the correlation between model performance and the size of the knowledge base constructed by the Reflective Exploration Module. 
As shown in Fig. \ref{fig_hint}, there is an improvement in both evaluation metrics as the size of the knowledge base increases to 20. 
The improvement is attributed to the diversity of the generated reflective hints by covering a broader range of common errors of LLM inference on SBR. 
However, when the knowledge base size becomes too large, there is a slight performance drop due to the increased complexity and difficulty in the retrieval agent optimization. 
Thus, we suggest adopting a moderate hint knowledge base size.

\captionsetup[figure]{belowskip=-10pt, aboveskip=1pt}
\begin{figure}[t] \centering
\vspace{-10pt}
 \includegraphics[width=7.7cm]{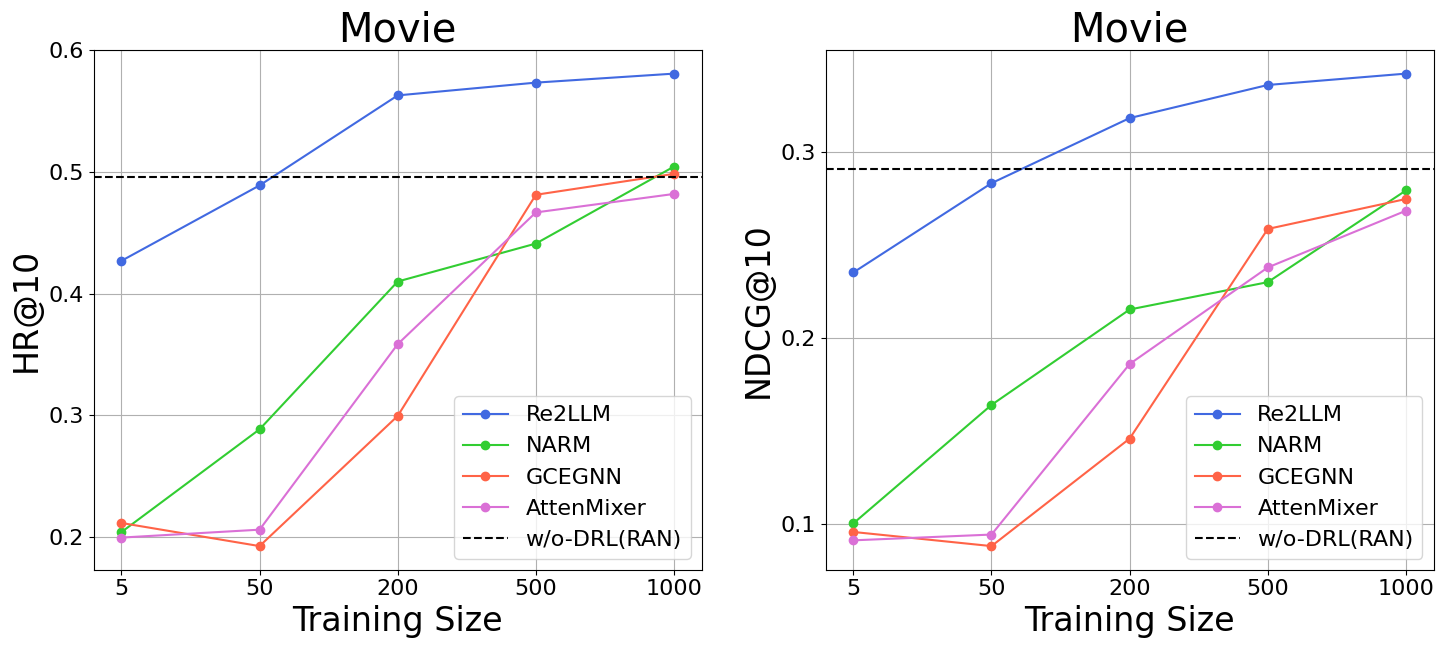}
\caption{Performance of the representative baselines and our method with varying few-shot training size for retrieval agent. The dotted lines indicate w/o-DRL(RAN) performance.}\label{fig_rl}
\vspace{-5pt}
\end{figure}

\subsubsection{Few-shot Training Size.}
We also study how model performance is affected by the number of few-shot training samples used for the retrieval model training. {Fig. \ref{fig_rl} shows the comparison of Re2LLM, its variant w/o-DRL(RAN), and representative baselines across various few-shot training sizes\footnote{We use previously mentioned data augmentation \cite{augment} for baselines. The results are from the Movie domain. Similar trends occur for the VideoGame domain.}.
\textbf{First}, using too few samples (e.g., less than 50) for retrieval agent training results in inferior performance, which is even lower than w/o-DRL(RAN) with random hint selection. This is because the retrieval agent is under-trained with only very few samples.
\textbf{Second}, as the number of training samples increases, the performance of our method increases consistently, showing the growth of the generalization ability of the retrieval agent. Baseline methods also improve, but remain inferior to our proposed method across various few-shot training sizes.
\textbf{Third}, we also observe the limited improvement of Re2LLM when the number of samples exceeds a threshold. For the balance between effectiveness and efficiency, 500 is an appropriate selection for the few-shot training size of Re2LLM.}

\subsubsection{Reward Design}
In Table \ref{tab:reward}, we report the experimental results using two distinct reward designs for the Reinforcement Retrieval Module: comparative reward and absolute reward. 
The comparative reward  $R= m(O^*(s)) - m(O(s)))$ measures the improvement achieved by the hint-enhanced prompt over the basic prompt. 
The absolute reward $R^{'} = m(O^*(s))$ evaluates the performance of the hint-enhanced prompt alone. 
{We find that the comparative reward yields superior performance compared to the absolute reward, as the comparative strategy focuses on improvement owing to the selected hint rather than its overall performance. As a result, we adopt the comparative reward for our Re2LLM.}

\begin{table}[htbp]
\vspace{-5pt}
  \centering
  \renewcommand{\arraystretch}{1}
  \fontsize{8}{8}\selectfont
  \caption{Performance of Re2LLM with two reward designs.}
  \vspace{-5pt}
  \label{tab:reward}
  \vspace{-0.1in}
\begin{tabular}{c|cc|cc}
\toprule
 \multirow{2}[3]{*}{\textbf{Reward}} & \multicolumn{2}{c|}{\textbf{Movie}} & \multicolumn{2}{c}{\textbf{Game}} \\ 
 \cmidrule{2-5}
  & \multicolumn{1}{c}{HR@10} & \multicolumn{1}{c|}{NDCG@10}  & \multicolumn{1}{c}{HR@10}  & \multicolumn{1}{c}{NDCG@10}\\ 
 \midrule
absolute    &0.5143&	0.3147&	0.7026	&0.4542\\
\midrule
{comparative} & \textbf{0.5735} & \textbf{0.3358} & \textbf{0.7173} &  \textbf{0.4689} \\
\bottomrule
\end{tabular}
\vspace{-10pt}
\end{table}

\subsection{Case Study (RQ2)}\label{casestudy} 
To further study the effectiveness of the retrieval agent, we compare Re2LLM, variant w/o-HE, and variant w/o-DRL(RAN) in two cases. 
Case 1 illustrates the scenario where the LLM backbone fails to hit the target item without a hint (i.e., the variant w/o-HE). Our trained retrieval agent in Re2LLM successfully retrieves the tailored hint and hits the target item at the third place of the top-10 recommendation list, demonstrating its ability to learn from task-specific feedback. However, the variant w/o-DRL(RAN) fails to replicate the correct prediction as a random hint may not be relevant to the session. Case 2 illustrates the scenario where the LLM backbone succeeds in hitting the target item without a hint (i.e., the variant w/o-HE). The trained agent in Re2LLM can further improve recommendation results by hitting the target item in a higher-ranking position (i.e., from the third place to the first place).
 On the contrary, a randomly selected hint by the variant w/o-DRL(RAN) shows a negative impact on the inference result, making the LLM miss the target item. Specifically, the session shows a preference for comedy, but the inappropriate hint instructs LLM's attention on the movie production years, thus misleading the LLM to generate incorrect predictions. 
 These two cases reveal the necessity of effectively exploring and efficiently utilizing specialized knowledge (i.e., hints) for LLMs in SBR, validating our motivation and the efficacy of key modules in our Re2LLM.

\definecolor{myboxcolor}{RGB}{0,100,0} 
\definecolor{myboxcolor2}{RGB}{100,0,0} 
\newtcolorbox{mybox}[1]{
  colback=myboxcolor!5!white,
  colframe=myboxcolor!75!black,
  fonttitle=\bfseries,
  title=#1,
  halign title=center, 
  halign=flush left,   
  left=0mm,
  right=0mm
}
\begin{mybox}{Case Study 1: \emph{Trained retriever}}
\vspace{-7pt}
\small{\textit{\textbf{Prompt}: I watched the following movies in order: \{Titanic; Leaving Las Vegas...\}. Based on these interactions, recommend a movie for me to watch next from a candidate set: \{\color{blue}$\mathcal{C}$\color{black}\}. Please recommend from the candidate set. List the top 10 recommendations in numbered bullet points. Hint: \{\color{blue}$h$\color{black}\}.
\newline 
\textbf{Answer by variant w/o-HE}, where \color{blue} $h$ \color{black} = `':
\newline  1. The Day After Tomorrow; 2. The Bourne Identity; 3. Last Man Standing; 4. The Last Castle; 5. Resident Evil: Extinction; 6. Deep Blue Sea; 7. True Grit; 8. Courage Under Fire; 9. Kiss of the Dragon; 10. The Wrestler
\newline 
\textbf{Answer by Re2LLM}, where \color{myboxcolor!75!black}{\textbf{retrieved hint}} \color{blue} $h$ \color{black} = `\textbf{\color{myboxcolor!75!black}{Consider films that originated from other media, such as books, radio, or TV shows}}':
\newline 1. The Bourne Identity; 2. True Grit;  
\textbf{\color{myboxcolor!75!black}{3. Aladdin}}; 4. The Parent Trap; 5. The Picture of Dorian Gray; 6. Breakfast at Tiffany's; 7. The Jazz Singer; 8. The Last Castle; 9. The Stepfather; 10. Cujo
\newline 
\textbf{Answer by w/o-DRL(RAN)}, where \color{myboxcolor2!75!black}{\textbf{random hint}} \color{blue} $h$ \color{black} = `\textbf{\color{myboxcolor2!75!black}{Pay attention to actors who appear in multiple movies from the watched list and consider other films featuring these actors}}':
\newline 1. The Day After Tomorrow; 2. Taken; 3. The Bourne Identity; 4. Last Man Standing; 5. The Last Castle; 6. The Wrestler; 7. Redbelt; 8. The Stepfather; 9. Kiss of the Dragon; 10. Courage Under Fire
}}
\vspace{-8pt}
\end{mybox} 
 
\newtcolorbox{mybox2}[1]{
  colback=myboxcolor2!5!white,
  colframe=myboxcolor2!75!black,
  fonttitle=\bfseries,
  title=#1,
  halign title=center, 
  halign=flush left,   
  left=0mm,
  right=0mm
}
\vspace{-10pt}
\begin{mybox2}{Case Study 2: \emph{Random retriever}}
\vspace{-7pt}
\small{\textit{\textbf{Prompt}: I watched the following movies in order: \{Xizao; Hero at Large; ...\} Based on these interactions, recommend a movie for me to watch next from a candidate set: \{\color{blue}$\mathcal{C}$\color{black}\}. Please recommend from the candidate set. List the top 10 recommendations in numbered bullet points. Hint: \{\color{blue}$h$\color{black}\}.
\newline 
\textbf{Answer by variant w/o-HE}, where \color{blue} $h$ \color{black} = `':\newline 1. Iceman; 2. Banlieue 13; \textbf{\color{myboxcolor!75!black}{3. Siu lam juk kau}}
; 4. Hot Fuzz; 5. King of New York; 6. Cidade dos Homens; 7. Blood Diamond; 8. Fantastic Four; 9. The Perfect Storm; 10. Chaplin
\newline
\textbf{Answer by Re2LLM}, where \color{myboxcolor!75!black}{\textbf{retrieved hint}} \color{blue} $h$ \color{black} = `\textbf{\color{myboxcolor!75!black}{Focus on films with a strong comedic element, particularly those that blend humor with other genres}}':
\newline \textbf{\color{myboxcolor!75!black}{1. Siu lam juk kau}}\color{black}; 2. Iceman; 3. Banlieue 13; 4. Hot Fuzz; 5. Broken Flowers; 6. The Perfect Storm; 7. Fantastic Four; 8. Blood Diamond; 9. King of New York; 10. Chaplin
\newline
\textbf{Answer by w/o-DRL(RAN)}, where \color{myboxcolor2!75!black}{\textbf{random hint}} \color{blue} $h$ \color{black} = `\textbf{\color{myboxcolor2!75!black}{Consider the release years of movies in the user’s history for era preference}}':
 \newline \color{black}{1. Blood Diamond; 2. Hot Fuzz; 3. Banlieue 13; 4. Freedom Writers; 5. Fantastic Four; 6. 2 Fast 2 Furious; 7. The Perfect Storm; 8. Iceman; 9. Ice Station Zebra; 10. Exit Wounds 2001}
}}
\vspace{-8pt}
\end{mybox2}
\vspace{-5pt}

\section{Conclusion}
In this paper, we propose Re2LLM, a Reflective Reinforcement LLM for SBR, aiming to improve performance by identifying and rectifying common errors in LLM inference. 
We present a novel learning paradigm that merges the capabilities of LLMs with adaptable model training procedures. 
Specifically, Re2LLM harnesses the self-reflection ability of LLMs to capture specialized knowledge and create a hint knowledge base with an automated filtering strategy. 
Then, trained via DRL, a lightweight retrieval model learns to select proper hints guided by task-specific feedback to facilitate session-aware inference for better recommendation results. 
We demonstrate the effectiveness of our method through extensive experiments in two real-world datasets across two evaluation settings. 
In addition, ablation studies and hyper-parameter analysis further validate our underlying motivations and designs. 
In future work, the retrieval model could be extended with more flexible functionalities, such as integrating hints and multi-modal contextual information.
\bibliographystyle{ACM-Reference-Format}
\bibliography{output.bib}


\begin{thebibliography}{60}


\ifx \showCODEN    \undefined \def \showCODEN     #1{\unskip}     \fi
\ifx \showDOI      \undefined \def \showDOI       #1{#1}\fi
\ifx \showISBNx    \undefined \def \showISBNx     #1{\unskip}     \fi
\ifx \showISBNxiii \undefined \def \showISBNxiii  #1{\unskip}     \fi
\ifx \showISSN     \undefined \def \showISSN      #1{\unskip}     \fi
\ifx \showLCCN     \undefined \def \showLCCN      #1{\unskip}     \fi
\ifx \shownote     \undefined \def \shownote      #1{#1}          \fi
\ifx \showarticletitle \undefined \def \showarticletitle #1{#1}   \fi
\ifx \showURL      \undefined \def \showURL       {\relax}        \fi
\providecommand\bibfield[2]{#2}
\providecommand\bibinfo[2]{#2}
\providecommand\natexlab[1]{#1}
\providecommand\showeprint[2][]{arXiv:#2}

\bibitem[Bao et~al\mbox{.}(2023)]%
        {tallrec}
\bibfield{author}{\bibinfo{person}{Keqin Bao}, \bibinfo{person}{Jizhi Zhang}, \bibinfo{person}{Yang Zhang}, \bibinfo{person}{Wenjie Wang}, \bibinfo{person}{Fuli Feng}, {and} \bibinfo{person}{Xiangnan He}.} \bibinfo{year}{2023}\natexlab{}.
\newblock \showarticletitle{TALLRec: An effective and efficient tuning framework to align large language model with recommendation}. In \bibinfo{booktitle}{\emph{Proceedings of the 17th ACM Conference on Recommender Systems (RecSys)}}. \bibinfo{pages}{1007–1014}.
\newblock


\bibitem[Chen et~al\mbox{.}(2022)]%
        {autogsr}
\bibfield{author}{\bibinfo{person}{Jingfan Chen}, \bibinfo{person}{Guanghui Zhu}, \bibinfo{person}{Haojun Hou}, \bibinfo{person}{Chunfeng Yuan}, {and} \bibinfo{person}{Yihua Huang}.} \bibinfo{year}{2022}\natexlab{}.
\newblock \showarticletitle{AutoGSR: Neural Architecture Search for Graph-based Session Recommendation}. In \bibinfo{booktitle}{\emph{Proceedings of the 45th International ACM SIGIR Conference on Research and Development in Information Retrieval}} \emph{(\bibinfo{series}{SIGIR '22})}. \bibinfo{pages}{1694–1704}.
\newblock


\bibitem[Chen et~al\mbox{.}({[n.\,d.]})]%
        {knowledge-enhance}
\bibfield{author}{\bibinfo{person}{Qian Chen}, \bibinfo{person}{Zhiqiang Guo}, \bibinfo{person}{Jianjun Li}, {and} \bibinfo{person}{Guohui Li}.} \bibinfo{year}{[n.\,d.]}\natexlab{}.
\newblock \showarticletitle{Knowledge-enhanced Multi-View Graph Neural Networks for Session-based Recommendation}. In \bibinfo{booktitle}{\emph{Proceedings of the 46th International ACM SIGIR Conference on Research and Development in Information Retrieval}} \emph{(\bibinfo{series}{SIGIR '23})}. \bibinfo{pages}{352–361}.
\newblock


\bibitem[Dai et~al\mbox{.}(2023)]%
        {chatgptrs}
\bibfield{author}{\bibinfo{person}{Sunhao Dai}, \bibinfo{person}{Ninglu Shao}, \bibinfo{person}{Haiyuan Zhao}, \bibinfo{person}{Weijie Yu}, \bibinfo{person}{Zihua Si}, \bibinfo{person}{Chen Xu}, \bibinfo{person}{Zhongxiang Sun}, \bibinfo{person}{Xiao Zhang}, {and} \bibinfo{person}{Jun Xu}.} \bibinfo{year}{2023}\natexlab{}.
\newblock \showarticletitle{Uncovering ChatGPT’s Capabilities in Recommender Systems}. In \bibinfo{booktitle}{\emph{Proceedings of the 17th ACM Conference on Recommender Systems (RecSys)}}. \bibinfo{pages}{1126–1132}.
\newblock


\bibitem[Devlin et~al\mbox{.}(2019)]%
        {bert}
\bibfield{author}{\bibinfo{person}{Jacob Devlin}, \bibinfo{person}{Ming-Wei Chang}, \bibinfo{person}{Kenton Lee}, {and} \bibinfo{person}{Kristina Toutanova}.} \bibinfo{year}{2019}\natexlab{}.
\newblock \showarticletitle{{BERT}: Pre-training of Deep Bidirectional Transformers for Language Understanding}. In \bibinfo{booktitle}{\emph{Proceedings of the 2019 Conference of the North {A}merican Chapter of the Association for Computational Linguistics: Human Language Technologies, Volume 1 (Long and Short Papers)}}. \bibinfo{publisher}{Association for Computational Linguistics}, \bibinfo{pages}{4171--4186}.
\newblock


\bibitem[Du et~al\mbox{.}(2023)]%
        {du2023enhancing}
\bibfield{author}{\bibinfo{person}{Yingpeng Du}, \bibinfo{person}{Di Luo}, \bibinfo{person}{Rui Yan}, \bibinfo{person}{Hongzhi Liu}, \bibinfo{person}{Yang Song}, \bibinfo{person}{Hengshu Zhu}, {and} \bibinfo{person}{Jie Zhang}.} \bibinfo{year}{2023}\natexlab{}.
\newblock \showarticletitle{Enhancing job recommendation through llm-based generative adversarial networks}.
\newblock \bibinfo{journal}{\emph{arXiv preprint arXiv:2307.10747}} (\bibinfo{year}{2023}).
\newblock


\bibitem[Du et~al\mbox{.}(2024)]%
        {du2024large}
\bibfield{author}{\bibinfo{person}{Yingpeng Du}, \bibinfo{person}{Ziyan Wang}, \bibinfo{person}{Zhu Sun}, \bibinfo{person}{Haoyan Chua}, \bibinfo{person}{Hongzhi Liu}, \bibinfo{person}{Zhonghai Wu}, \bibinfo{person}{Yining Ma}, \bibinfo{person}{Jie Zhang}, {and} \bibinfo{person}{Youchen Sun}.} \bibinfo{year}{2024}\natexlab{}.
\newblock \showarticletitle{Large language model with graph convolution for recommendation}.
\newblock \bibinfo{journal}{\emph{arXiv preprint arXiv:2402.08859}} (\bibinfo{year}{2024}).
\newblock


\bibitem[Gao et~al\mbox{.}(2023)]%
        {chatrec}
\bibfield{author}{\bibinfo{person}{Yunfan Gao}, \bibinfo{person}{Tao Sheng}, \bibinfo{person}{Youlin Xiang}, \bibinfo{person}{Yun Xiong}, \bibinfo{person}{Haofen Wang}, {and} \bibinfo{person}{Jiawei Zhang}.} \bibinfo{year}{2023}\natexlab{}.
\newblock \bibinfo{title}{Chat-REC: Towards Interactive and Explainable LLMs-Augmented Recommender System}.
\newblock
\newblock
\showeprint[arxiv]{2303.14524}~[cs.IR]


\bibitem[Geng et~al\mbox{.}(2022)]%
        {p5}
\bibfield{author}{\bibinfo{person}{Shijie Geng}, \bibinfo{person}{Shuchang Liu}, \bibinfo{person}{Zuohui Fu}, \bibinfo{person}{Yingqiang Ge}, {and} \bibinfo{person}{Yongfeng Zhang}.} \bibinfo{year}{2022}\natexlab{}.
\newblock \showarticletitle{Recommendation as language processing (RLP): a unified pretrain, personalized prompt \& predict paradigm (P5)}. In \bibinfo{booktitle}{\emph{Proceedings of the 16th ACM Conference on Recommender Systems (RecSys)}}. \bibinfo{pages}{299–315}.
\newblock


\bibitem[Han et~al\mbox{.}(2022)]%
        {intro3}
\bibfield{author}{\bibinfo{person}{Qilong Han}, \bibinfo{person}{Chi Zhang}, \bibinfo{person}{Rui Chen}, \bibinfo{person}{Riwei Lai}, \bibinfo{person}{Hongtao Song}, {and} \bibinfo{person}{Li Li}.} \bibinfo{year}{2022}\natexlab{}.
\newblock \showarticletitle{Multi-Faceted Global Item Relation Learning for Session-Based Recommendation}. In \bibinfo{booktitle}{\emph{Proceedings of the 45th International ACM SIGIR Conference on Research and Development in Information Retrieval}} \emph{(\bibinfo{series}{SIGIR '22})}. \bibinfo{pages}{1705–1715}.
\newblock


\bibitem[He et~al\mbox{.}({[n.\,d.]})]%
        {llmconversation}
\bibfield{author}{\bibinfo{person}{Zhankui He}, \bibinfo{person}{Zhouhang Xie}, \bibinfo{person}{Rahul Jha}, \bibinfo{person}{Harald Steck}, \bibinfo{person}{Dawen Liang}, \bibinfo{person}{Yesu Feng}, \bibinfo{person}{Bodhisattwa~Prasad Majumder}, \bibinfo{person}{Nathan Kallus}, {and} \bibinfo{person}{Julian Mcauley}.} \bibinfo{year}{[n.\,d.]}\natexlab{}.
\newblock \showarticletitle{Large Language Models as Zero-Shot Conversational Recommenders}. In \bibinfo{booktitle}{\emph{Proceedings of the 32nd ACM International Conference on Information and Knowledge Management}} \emph{(\bibinfo{series}{CIKM '23})}. \bibinfo{pages}{720–730}.
\newblock


\bibitem[Hidasi and Karatzoglou(2018)]%
        {lossfuc}
\bibfield{author}{\bibinfo{person}{Bal{\'{a}}zs Hidasi} {and} \bibinfo{person}{Alexandros Karatzoglou}.} \bibinfo{year}{2018}\natexlab{}.
\newblock \showarticletitle{Recurrent Neural Networks with Top-k Gains for Session-based Recommendations}. In \bibinfo{booktitle}{\emph{Proceedings of the 27th {ACM} International Conference on Information and Knowledge Management, {CIKM} 2018, Torino, Italy, October 22-26, 2018}}. \bibinfo{publisher}{{ACM}}, \bibinfo{pages}{843--852}.
\newblock


\bibitem[Hidasi et~al\mbox{.}({[n.\,d.]})]%
        {DBLP:journals/corr/HidasiKBT15}
\bibfield{author}{\bibinfo{person}{Bal{\'{a}}zs Hidasi}, \bibinfo{person}{Alexandros Karatzoglou}, \bibinfo{person}{Linas Baltrunas}, {and} \bibinfo{person}{Domonkos Tikk}.} \bibinfo{year}{[n.\,d.]}\natexlab{}.
\newblock \showarticletitle{Session-based recommendations with recurrent neural networks}. In \bibinfo{booktitle}{\emph{4th International Conference on Learning Representations, {ICLR} 2016}}.
\newblock


\bibitem[Hou et~al\mbox{.}(2022)]%
        {intro2}
\bibfield{author}{\bibinfo{person}{Yupeng Hou}, \bibinfo{person}{Binbin Hu}, \bibinfo{person}{Zhiqiang Zhang}, {and} \bibinfo{person}{Wayne~Xin Zhao}.} \bibinfo{year}{2022}\natexlab{}.
\newblock \showarticletitle{CORE: Simple and Effective Session-Based Recommendation within Consistent Representation Space} \emph{(\bibinfo{series}{SIGIR '22})}. \bibinfo{pages}{1796–1801}.
\newblock


\bibitem[Hou et~al\mbox{.}({[n.\,d.]})]%
        {textsequence}
\bibfield{author}{\bibinfo{person}{Yupeng Hou}, \bibinfo{person}{Shanlei Mu}, \bibinfo{person}{Wayne~Xin Zhao}, \bibinfo{person}{Yaliang Li}, \bibinfo{person}{Bolin Ding}, {and} \bibinfo{person}{Ji-Rong Wen}.} \bibinfo{year}{[n.\,d.]}\natexlab{}.
\newblock \showarticletitle{Towards Universal Sequence Representation Learning for Recommender Systems}. In \bibinfo{booktitle}{\emph{Proceedings of the 28th ACM SIGKDD Conference on Knowledge Discovery and Data Mining}} \emph{(\bibinfo{series}{KDD '22})}. \bibinfo{pages}{585–593}.
\newblock


\bibitem[Hou et~al\mbox{.}(2023)]%
        {llmrank}
\bibfield{author}{\bibinfo{person}{Yupeng Hou}, \bibinfo{person}{Junjie Zhang}, \bibinfo{person}{Zihan Lin}, \bibinfo{person}{Hongyu Lu}, \bibinfo{person}{Ruobing Xie}, \bibinfo{person}{Julian McAuley}, {and} \bibinfo{person}{Wayne~Xin Zhao}.} \bibinfo{year}{2023}\natexlab{}.
\newblock \bibinfo{title}{Large language models are zero-shot rankers for recommender systems}.
\newblock
\newblock
\showeprint[arxiv]{2305.08845}


\bibitem[Hu et~al\mbox{.}(2022)]%
        {lora}
\bibfield{author}{\bibinfo{person}{Edward~J Hu}, \bibinfo{person}{Yelong Shen}, \bibinfo{person}{Phillip Wallis}, \bibinfo{person}{Zeyuan Allen-Zhu}, \bibinfo{person}{Yuanzhi Li}, \bibinfo{person}{Shean Wang}, \bibinfo{person}{Lu Wang}, {and} \bibinfo{person}{Weizhu Chen}.} \bibinfo{year}{2022}\natexlab{}.
\newblock \showarticletitle{Lo{RA}: Low-rank adaptation of large language models}. In \bibinfo{booktitle}{\emph{International Conference on Learning Representations (ICLR)}}.
\newblock


\bibitem[Ji et~al\mbox{.}(2023)]%
        {genrec}
\bibfield{author}{\bibinfo{person}{Jianchao Ji}, \bibinfo{person}{Zelong Li}, \bibinfo{person}{Shuyuan Xu}, \bibinfo{person}{Wenyue Hua}, \bibinfo{person}{Yingqiang Ge}, \bibinfo{person}{Juntao Tan}, {and} \bibinfo{person}{Yongfeng Zhang}.} \bibinfo{year}{2023}\natexlab{}.
\newblock \bibinfo{title}{GenRec: Large Language Model for Generative Recommendation}.
\newblock
\newblock


\bibitem[Kang et~al\mbox{.}(2023)]%
        {llmunderstand}
\bibfield{author}{\bibinfo{person}{Wang-Cheng Kang}, \bibinfo{person}{Jianmo Ni}, \bibinfo{person}{Nikhil Mehta}, \bibinfo{person}{Maheswaran Sathiamoorthy}, \bibinfo{person}{Lichan Hong}, \bibinfo{person}{Ed Chi}, {and} \bibinfo{person}{Derek~Zhiyuan Cheng}.} \bibinfo{year}{2023}\natexlab{}.
\newblock \bibinfo{title}{Do LLMs Understand User Preferences? Evaluating LLMs On User Rating Prediction}.
\newblock
\newblock
\showeprint[arxiv]{2305.06474}


\bibitem[Koren et~al\mbox{.}(2009)]%
        {koren2009matrix}
\bibfield{author}{\bibinfo{person}{Yehuda Koren}, \bibinfo{person}{Robert Bell}, {and} \bibinfo{person}{Chris Volinsky}.} \bibinfo{year}{2009}\natexlab{}.
\newblock \showarticletitle{Matrix factorization techniques for recommender systems}.
\newblock \bibinfo{journal}{\emph{Computer}}  \bibinfo{volume}{42} (\bibinfo{year}{2009}), \bibinfo{pages}{30--37}.
\newblock


\bibitem[Lai et~al\mbox{.}(2022)]%
        {intro1}
\bibfield{author}{\bibinfo{person}{Siqi Lai}, \bibinfo{person}{Erli Meng}, \bibinfo{person}{Fan Zhang}, \bibinfo{person}{Chenliang Li}, \bibinfo{person}{Bin Wang}, {and} \bibinfo{person}{Aixin Sun}.} \bibinfo{year}{2022}\natexlab{}.
\newblock \showarticletitle{An Attribute-Driven Mirror Graph Network for Session-Based Recommendation} \emph{(\bibinfo{series}{SIGIR '22})}. \bibinfo{pages}{1674–1683}.
\newblock


\bibitem[Li et~al\mbox{.}(2017)]%
        {narm}
\bibfield{author}{\bibinfo{person}{Jing Li}, \bibinfo{person}{Pengjie Ren}, \bibinfo{person}{Zhumin Chen}, \bibinfo{person}{Zhaochun Ren}, \bibinfo{person}{Tao Lian}, {and} \bibinfo{person}{Jun Ma}.} \bibinfo{year}{2017}\natexlab{}.
\newblock \showarticletitle{Neural attentive session-based recommendation}. In \bibinfo{booktitle}{\emph{Proceedings of the 2017 ACM on Conference on Information and Knowledge Management (CIKM)}}. \bibinfo{pages}{1419–1428}.
\newblock


\bibitem[Li et~al\mbox{.}(2023)]%
        {Li2023MultiModalityIA}
\bibfield{author}{\bibinfo{person}{Youhua Li}, \bibinfo{person}{Hanwen Du}, \bibinfo{person}{Yongxin Ni}, \bibinfo{person}{Pengpeng Zhao}, \bibinfo{person}{Qi Guo}, \bibinfo{person}{Fajie Yuan}, {and} \bibinfo{person}{Xiaofang Zhou}.} \bibinfo{year}{2023}\natexlab{}.
\newblock \showarticletitle{Multi-modality is all you need for transferable recommender systems}.
\newblock \bibinfo{journal}{\emph{ArXiv}}  \bibinfo{volume}{abs/2312.09602} (\bibinfo{year}{2023}).
\newblock


\bibitem[Li et~al\mbox{.}(2022)]%
        {hide}
\bibfield{author}{\bibinfo{person}{Yinfeng Li}, \bibinfo{person}{Chen Gao}, \bibinfo{person}{Hengliang Luo}, \bibinfo{person}{Depeng Jin}, {and} \bibinfo{person}{Yong Li}.} \bibinfo{year}{2022}\natexlab{}.
\newblock \showarticletitle{Enhancing hypergraph neural networks with intent disentanglement for session-based recommendation}. In \bibinfo{booktitle}{\emph{Proceedings of the 45th International ACM SIGIR Conference on Research and Development in Information Retrieval}}. \bibinfo{pages}{1997–2002}.
\newblock


\bibitem[Liu et~al\mbox{.}(2024)]%
        {once}
\bibfield{author}{\bibinfo{person}{Qijiong Liu}, \bibinfo{person}{Nuo Chen}, \bibinfo{person}{Tetsuya Sakai}, {and} \bibinfo{person}{Xiao-Ming Wu}.} \bibinfo{year}{2024}\natexlab{}.
\newblock \showarticletitle{ONCE: Boosting Content-based Recommendation with Both Open- and Closed-source Large Language Models}. In \bibinfo{booktitle}{\emph{Proceedings of the Seventeen ACM International Conference on Web Search and Data Mining}}.
\newblock


\bibitem[Liu et~al\mbox{.}(2018)]%
        {stamp}
\bibfield{author}{\bibinfo{person}{Qiao Liu}, \bibinfo{person}{Yifu Zeng}, \bibinfo{person}{Refuoe Mokhosi}, {and} \bibinfo{person}{Haibin Zhang}.} \bibinfo{year}{2018}\natexlab{}.
\newblock \showarticletitle{STAMP: Short-term attention/memory priority model for session-based recommendation.}. In \bibinfo{booktitle}{\emph{KDD}}. \bibinfo{publisher}{ACM}, \bibinfo{pages}{1831--1839}.
\newblock


\bibitem[Madaan et~al\mbox{.}(2023)]%
        {selfrefine}
\bibfield{author}{\bibinfo{person}{Aman Madaan}, \bibinfo{person}{Niket Tandon}, \bibinfo{person}{Prakhar Gupta}, \bibinfo{person}{Skyler Hallinan}, \bibinfo{person}{Luyu Gao}, \bibinfo{person}{Sarah Wiegreffe}, \bibinfo{person}{Uri Alon}, \bibinfo{person}{Nouha Dziri}, \bibinfo{person}{Shrimai Prabhumoye}, \bibinfo{person}{Yiming Yang}, \bibinfo{person}{Sean Welleck}, \bibinfo{person}{Bodhisattwa~Prasad Majumder}, \bibinfo{person}{Shashank Gupta}, \bibinfo{person}{Amir Yazdanbakhsh}, {and} \bibinfo{person}{Peter Clark}.} \bibinfo{year}{2023}\natexlab{}.
\newblock \bibinfo{title}{Self-Refine: Iterative Refinement with Self-Feedback}.
\newblock
\newblock
\showeprint[arxiv]{2303.17651}~[cs.CL]


\bibitem[Meng et~al\mbox{.}(2020)]%
        {coldstart}
\bibfield{author}{\bibinfo{person}{Wenjing Meng}, \bibinfo{person}{Deqing Yang}, {and} \bibinfo{person}{Yanghua Xiao}.} \bibinfo{year}{2020}\natexlab{}.
\newblock \showarticletitle{Incorporating User Micro-Behaviors and Item Knowledge into Multi-Task Learning for Session-Based Recommendation}. In \bibinfo{booktitle}{\emph{Proceedings of the 43rd International ACM SIGIR Conference on Research and Development in Information Retrieval}} \emph{(\bibinfo{series}{SIGIR '20})}. \bibinfo{pages}{1091–1100}.
\newblock


\bibitem[Pan et~al\mbox{.}(2023)]%
        {correctionsurvey}
\bibfield{author}{\bibinfo{person}{Liangming Pan}, \bibinfo{person}{Michael Saxon}, \bibinfo{person}{Wenda Xu}, \bibinfo{person}{Deepak Nathani}, \bibinfo{person}{Xinyi Wang}, {and} \bibinfo{person}{William~Yang Wang}.} \bibinfo{year}{2023}\natexlab{}.
\newblock \bibinfo{title}{Automatically Correcting Large Language Models: Surveying the landscape of diverse self-correction strategies}.
\newblock
\newblock
\showeprint[arxiv]{2308.03188}~[cs.CL]


\bibitem[Pan et~al\mbox{.}(2020)]%
        {star}
\bibfield{author}{\bibinfo{person}{Zhiqiang Pan}, \bibinfo{person}{Fei Cai}, \bibinfo{person}{Wanyu Chen}, \bibinfo{person}{Honghui Chen}, {and} \bibinfo{person}{Maarten de Rijke}.} \bibinfo{year}{2020}\natexlab{}.
\newblock \showarticletitle{Star graph neural networks for session-based recommendation}. In \bibinfo{booktitle}{\emph{Proceedings of the 29th ACM International Conference on Information \& Knowledge Management (CIKM)}}. \bibinfo{pages}{1195–1204}.
\newblock


\bibitem[Pryzant et~al\mbox{.}(2023)]%
        {promptopt}
\bibfield{author}{\bibinfo{person}{Reid Pryzant}, \bibinfo{person}{Dan Iter}, \bibinfo{person}{Jerry Li}, \bibinfo{person}{Yin Lee}, \bibinfo{person}{Chenguang Zhu}, {and} \bibinfo{person}{Michael Zeng}.} \bibinfo{year}{2023}\natexlab{}.
\newblock \showarticletitle{Automatic Prompt Optimization with {``}Gradient Descent{''} and Beam Search}. In \bibinfo{booktitle}{\emph{Proceedings of the 2023 Conference on Empirical Methods in Natural Language Processing}}. \bibinfo{publisher}{Association for Computational Linguistics}, \bibinfo{address}{Singapore}, \bibinfo{pages}{7957--7968}.
\newblock


\bibitem[Rendle et~al\mbox{.}(2010)]%
        {fpmc}
\bibfield{author}{\bibinfo{person}{Steffen Rendle}, \bibinfo{person}{Christoph Freudenthaler}, {and} \bibinfo{person}{Lars Schmidt-Thieme}.} \bibinfo{year}{2010}\natexlab{}.
\newblock \showarticletitle{Factorizing personalized Markov chains for next-basket recommendation}. In \bibinfo{booktitle}{\emph{Proceedings of the 19th International Conference on World Wide Web}} \emph{(\bibinfo{series}{WWW '10})}. \bibinfo{pages}{811–820}.
\newblock


\bibitem[Schulman et~al\mbox{.}(2017)]%
        {ppo}
\bibfield{author}{\bibinfo{person}{John Schulman}, \bibinfo{person}{Filip Wolski}, \bibinfo{person}{Prafulla Dhariwal}, \bibinfo{person}{Alec Radford}, {and} \bibinfo{person}{Oleg Klimov}.} \bibinfo{year}{2017}\natexlab{}.
\newblock \bibinfo{title}{Proximal Policy Optimization Algorithms}.
\newblock
\newblock
\showeprint[arxiv]{1707.06347}~[cs.LG]


\bibitem[Shinn et~al\mbox{.}(2023)]%
        {reflexion}
\bibfield{author}{\bibinfo{person}{Noah Shinn}, \bibinfo{person}{Federico Cassano}, \bibinfo{person}{Edward Berman}, \bibinfo{person}{Ashwin Gopinath}, \bibinfo{person}{Karthik Narasimhan}, {and} \bibinfo{person}{Shunyu Yao}.} \bibinfo{year}{2023}\natexlab{}.
\newblock \bibinfo{title}{Reflexion: Language Agents with Verbal Reinforcement Learning}.
\newblock
\newblock
\showeprint[arxiv]{2303.11366}~[cs.AI]


\bibitem[Sun et~al\mbox{.}(2022)]%
        {daisyrec2}
\bibfield{author}{\bibinfo{person}{Zhu Sun}, \bibinfo{person}{Hui Fang}, \bibinfo{person}{Jie Yang}, \bibinfo{person}{Xinghua Qu}, \bibinfo{person}{Hongyang Liu}, \bibinfo{person}{Di Yu}, \bibinfo{person}{Yew-Soon Ong}, {and} \bibinfo{person}{Jie Zhang}.} \bibinfo{year}{2022}\natexlab{}.
\newblock \showarticletitle{DaisyRec 2.0: Benchmarking Recommendation for Rigorous Evaluation}.
\newblock \bibinfo{journal}{\emph{IEEE Transactions on Pattern Analysis and Machine Intelligence (TPAMI)}} (\bibinfo{year}{2022}).
\newblock


\bibitem[Sun et~al\mbox{.}(2019)]%
        {sun2019research}
\bibfield{author}{\bibinfo{person}{Zhu Sun}, \bibinfo{person}{Qing Guo}, \bibinfo{person}{Jie Yang}, \bibinfo{person}{Hui Fang}, \bibinfo{person}{Guibing Guo}, \bibinfo{person}{Jie Zhang}, {and} \bibinfo{person}{Robin Burke}.} \bibinfo{year}{2019}\natexlab{}.
\newblock \showarticletitle{Research commentary on recommendations with side information: A survey and research directions}.
\newblock \bibinfo{journal}{\emph{Electronic Commerce Research and Applications}}  \bibinfo{volume}{37} (\bibinfo{year}{2019}), \bibinfo{pages}{100879}.
\newblock


\bibitem[Sun et~al\mbox{.}(2023)]%
        {sun2023large}
\bibfield{author}{\bibinfo{person}{Zhu Sun}, \bibinfo{person}{Hongyang Liu}, \bibinfo{person}{Xinghua Qu}, \bibinfo{person}{Kaidong Feng}, \bibinfo{person}{Yan Wang}, {and} \bibinfo{person}{Yew-Soon Ong}.} \bibinfo{year}{2023}\natexlab{}.
\newblock \bibinfo{title}{Large Language Models for Intent-Driven Session Recommendations}.
\newblock
\newblock
\showeprint[arxiv]{2312.07552}~[cs.CL]


\bibitem[Sun et~al\mbox{.}(2020)]%
        {daisyrec}
\bibfield{author}{\bibinfo{person}{Zhu Sun}, \bibinfo{person}{Di Yu}, \bibinfo{person}{Hui Fang}, \bibinfo{person}{Jie Yang}, \bibinfo{person}{Xinghua Qu}, \bibinfo{person}{Jie Zhang}, {and} \bibinfo{person}{Cong Geng}.} \bibinfo{year}{2020}\natexlab{}.
\newblock \showarticletitle{Are We Evaluating Rigorously? Benchmarking Recommendation for Reproducible Evaluation and Fair Comparison}. In \bibinfo{booktitle}{\emph{Proceedings of the 14th ACM Conference on Recommender Systems}}.
\newblock


\bibitem[Tan et~al\mbox{.}(2016)]%
        {augment}
\bibfield{author}{\bibinfo{person}{Yong~Kiam Tan}, \bibinfo{person}{Xinxing Xu}, {and} \bibinfo{person}{Yong Liu}.} \bibinfo{year}{2016}\natexlab{}.
\newblock \showarticletitle{Improved recurrent neural networks for session-based recommendations}. In \bibinfo{booktitle}{\emph{Proceedings of the 1st Workshop on Deep Learning for Recommender Systems}}. \bibinfo{publisher}{Association for Computing Machinery}, \bibinfo{pages}{17–22}.
\newblock


\bibitem[Touvron et~al\mbox{.}(2023)]%
        {llama}
\bibfield{author}{\bibinfo{person}{Hugo Touvron}, \bibinfo{person}{Thibaut Lavril}, \bibinfo{person}{Gautier Izacard}, \bibinfo{person}{Xavier Martinet}, \bibinfo{person}{Marie-Anne Lachaux}, \bibinfo{person}{Timothée Lacroix}, \bibinfo{person}{Baptiste Rozière}, \bibinfo{person}{Naman Goyal}, \bibinfo{person}{Eric Hambro}, \bibinfo{person}{Faisal Azhar}, \bibinfo{person}{Aurelien Rodriguez}, \bibinfo{person}{Armand Joulin}, \bibinfo{person}{Edouard Grave}, {and} \bibinfo{person}{Guillaume Lample}.} \bibinfo{year}{2023}\natexlab{}.
\newblock \bibinfo{title}{LLaMA: Open and Efficient Foundation Language Models}.
\newblock
\newblock
\showeprint[arxiv]{2302.13971}


\bibitem[Vaswani et~al\mbox{.}(2017)]%
        {attentionalluneed}
\bibfield{author}{\bibinfo{person}{Ashish Vaswani}, \bibinfo{person}{Noam Shazeer}, \bibinfo{person}{Niki Parmar}, \bibinfo{person}{Jakob Uszkoreit}, \bibinfo{person}{Llion Jones}, \bibinfo{person}{Aidan~N Gomez}, \bibinfo{person}{\L~ukasz Kaiser}, {and} \bibinfo{person}{Illia Polosukhin}.} \bibinfo{year}{2017}\natexlab{}.
\newblock \showarticletitle{Attention is All you Need}. In \bibinfo{booktitle}{\emph{Advances in Neural Information Processing Systems}}, \bibfield{editor}{\bibinfo{person}{I.~Guyon}, \bibinfo{person}{U.~Von Luxburg}, \bibinfo{person}{S.~Bengio}, \bibinfo{person}{H.~Wallach}, \bibinfo{person}{R.~Fergus}, \bibinfo{person}{S.~Vishwanathan}, {and} \bibinfo{person}{R.~Garnett}} (Eds.), Vol.~\bibinfo{volume}{30}. \bibinfo{publisher}{Curran Associates, Inc.}
\newblock


\bibitem[Wang and Lim(2023)]%
        {nir}
\bibfield{author}{\bibinfo{person}{Lei Wang} {and} \bibinfo{person}{Ee-Peng Lim}.} \bibinfo{year}{2023}\natexlab{}.
\newblock \bibinfo{title}{Zero-Shot Next-Item Recommendation using Large Pretrained Language Models}.
\newblock
\newblock
\showeprint[arxiv]{2304.03153}


\bibitem[Wang et~al\mbox{.}(2019)]%
        {mcprn}
\bibfield{author}{\bibinfo{person}{Shoujin Wang}, \bibinfo{person}{Liang Hu}, \bibinfo{person}{Yan Wang}, \bibinfo{person}{Quan~Z. Sheng}, \bibinfo{person}{Mehmet Orgun}, {and} \bibinfo{person}{Longbing Cao}.} \bibinfo{year}{2019}\natexlab{}.
\newblock \showarticletitle{Modeling multi-purpose sessions for next-item recommendations via mixture-channel purpose routing networks}. In \bibinfo{booktitle}{\emph{Proceedings of the Twenty-Eighth International Joint Conference on Artificial Intelligence, {IJCAI-19}}}. \bibinfo{pages}{3771--3777}.
\newblock


\bibitem[Wang et~al\mbox{.}(2023)]%
        {drdt}
\bibfield{author}{\bibinfo{person}{Yu Wang}, \bibinfo{person}{Zhiwei Liu}, \bibinfo{person}{Jianguo Zhang}, \bibinfo{person}{Weiran Yao}, \bibinfo{person}{Shelby Heinecke}, {and} \bibinfo{person}{Philip~S. Yu}.} \bibinfo{year}{2023}\natexlab{}.
\newblock \bibinfo{title}{DRDT: Dynamic Reflection with Divergent Thinking for LLM-based Sequential Recommendation}.
\newblock
\newblock
\showeprint[arxiv]{2312.11336}


\bibitem[Wang et~al\mbox{.}(2020)]%
        {gcegnn}
\bibfield{author}{\bibinfo{person}{Ziyang Wang}, \bibinfo{person}{Wei Wei}, \bibinfo{person}{Gao Cong}, \bibinfo{person}{Xiao-Li Li}, \bibinfo{person}{Xian-Ling Mao}, {and} \bibinfo{person}{Minghui Qiu}.} \bibinfo{year}{2020}\natexlab{}.
\newblock \showarticletitle{Global context enhanced graph neural networks for session-based recommendation}. In \bibinfo{booktitle}{\emph{Proceedings of the 43rd International ACM SIGIR Conference on Research and Development in Information Retrieval}}. \bibinfo{pages}{169–178}.
\newblock


\bibitem[Wei et~al\mbox{.}(2022)]%
        {cot}
\bibfield{author}{\bibinfo{person}{Jason Wei}, \bibinfo{person}{Xuezhi Wang}, \bibinfo{person}{Dale Schuurmans}, \bibinfo{person}{Maarten Bosma}, \bibinfo{person}{brian ichter}, \bibinfo{person}{Fei Xia}, \bibinfo{person}{Ed Chi}, \bibinfo{person}{Quoc~V Le}, {and} \bibinfo{person}{Denny Zhou}.} \bibinfo{year}{2022}\natexlab{}.
\newblock \showarticletitle{Chain-of-Thought Prompting Elicits Reasoning in Large Language Models}. In \bibinfo{booktitle}{\emph{Advances in Neural Information Processing Systems}}, Vol.~\bibinfo{volume}{35}. \bibinfo{publisher}{Curran Associates, Inc.}, \bibinfo{pages}{24824--24837}.
\newblock


\bibitem[Wu et~al\mbox{.}(2019)]%
        {srgnn}
\bibfield{author}{\bibinfo{person}{Shu Wu}, \bibinfo{person}{Yuyuan Tang}, \bibinfo{person}{Yanqiao Zhu}, \bibinfo{person}{Liang Wang}, \bibinfo{person}{Xing Xie}, {and} \bibinfo{person}{Tieniu Tan}.} \bibinfo{year}{2019}\natexlab{}.
\newblock \showarticletitle{Session-based recommendation with graph neural networks}. In \bibinfo{booktitle}{\emph{Proceedings of the Thirty-Third AAAI Conference}}.
\newblock


\bibitem[Xie et~al\mbox{.}(2022)]%
        {decouplesideinfo}
\bibfield{author}{\bibinfo{person}{Yueqi Xie}, \bibinfo{person}{Peilin Zhou}, {and} \bibinfo{person}{Sunghun Kim}.} \bibinfo{year}{2022}\natexlab{}.
\newblock \showarticletitle{Decoupled Side Information Fusion for Sequential Recommendation}. In \bibinfo{booktitle}{\emph{Proceedings of the 45th International ACM SIGIR Conference on Research and Development in Information Retrieval}} \emph{(\bibinfo{series}{SIGIR '22})}. \bibinfo{pages}{1611–1621}.
\newblock


\bibitem[Xu et~al\mbox{.}(2019)]%
        {gnn2}
\bibfield{author}{\bibinfo{person}{Chengfeng Xu}, \bibinfo{person}{Pengpeng Zhao}, \bibinfo{person}{Yanchi Liu}, \bibinfo{person}{Victor~S. Sheng}, \bibinfo{person}{Jiajie Xu}, \bibinfo{person}{Fuzhen Zhuang}, \bibinfo{person}{Junhua Fang}, {and} \bibinfo{person}{Xiaofang Zhou}.} \bibinfo{year}{2019}\natexlab{}.
\newblock \showarticletitle{Graph contextualized self-attention network for session-based recommendation}. In \bibinfo{booktitle}{\emph{Proceedings of the 28th International Joint Conference on Artificial Intelligence (IJCAI)}}. \bibinfo{pages}{3940–3946}.
\newblock


\bibitem[Yang et~al\mbox{.}(2023a)]%
        {palr}
\bibfield{author}{\bibinfo{person}{Fan Yang}, \bibinfo{person}{Zheng Chen}, \bibinfo{person}{Ziyan Jiang}, \bibinfo{person}{Eunah Cho}, \bibinfo{person}{Xiaojiang Huang}, {and} \bibinfo{person}{Yanbin Lu}.} \bibinfo{year}{2023}\natexlab{a}.
\newblock \bibinfo{title}{PALR: Personalization aware LLMs for recommendation}.
\newblock
\newblock
\showeprint[arxiv]{2305.07622}


\bibitem[Yang et~al\mbox{.}(2023b)]%
        {loam}
\bibfield{author}{\bibinfo{person}{Heeyoon Yang}, \bibinfo{person}{YunSeok Choi}, \bibinfo{person}{Gahyung Kim}, {and} \bibinfo{person}{Jee-Hyong Lee}.} \bibinfo{year}{2023}\natexlab{b}.
\newblock \showarticletitle{LOAM: Improving Long-tail Session-based Recommendation via Niche Walk Augmentation and Tail Session Mixup} \emph{(\bibinfo{series}{SIGIR '23})}. \bibinfo{pages}{527–536}.
\newblock


\bibitem[Yao et~al\mbox{.}(2023b)]%
        {tot}
\bibfield{author}{\bibinfo{person}{Shunyu Yao}, \bibinfo{person}{Dian Yu}, \bibinfo{person}{Jeffrey Zhao}, \bibinfo{person}{Izhak Shafran}, \bibinfo{person}{Thomas~L. Griffiths}, \bibinfo{person}{Yuan Cao}, {and} \bibinfo{person}{Karthik Narasimhan}.} \bibinfo{year}{2023}\natexlab{b}.
\newblock \bibinfo{title}{{Tree of Thoughts}: Deliberate Problem Solving with Large Language Models}.
\newblock
\newblock
\showeprint[arxiv]{2305.10601}~[cs.CL]


\bibitem[Yao et~al\mbox{.}(2023a)]%
        {retroformer}
\bibfield{author}{\bibinfo{person}{Weiran Yao}, \bibinfo{person}{Shelby Heinecke}, \bibinfo{person}{Juan~Carlos Niebles}, \bibinfo{person}{Zhiwei Liu}, \bibinfo{person}{Yihao Feng}, \bibinfo{person}{Le Xue}, \bibinfo{person}{Rithesh Murthy}, \bibinfo{person}{Zeyuan Chen}, \bibinfo{person}{Jianguo Zhang}, \bibinfo{person}{Devansh Arpit}, \bibinfo{person}{Ran Xu}, \bibinfo{person}{Phil Mui}, \bibinfo{person}{Huan Wang}, \bibinfo{person}{Caiming Xiong}, {and} \bibinfo{person}{Silvio Savarese}.} \bibinfo{year}{2023}\natexlab{a}.
\newblock \bibinfo{title}{Retroformer: Retrospective Large Language Agents with Policy Gradient Optimization}.
\newblock
\newblock
\showeprint[arxiv]{2308.02151}~[cs.CL]


\bibitem[Yu et~al\mbox{.}(2020)]%
        {tagnn}
\bibfield{author}{\bibinfo{person}{Feng Yu}, \bibinfo{person}{Yanqiao Zhu}, \bibinfo{person}{Qiang Liu}, \bibinfo{person}{Shu Wu}, \bibinfo{person}{Liang Wang}, {and} \bibinfo{person}{Tieniu Tan}.} \bibinfo{year}{2020}\natexlab{}.
\newblock \showarticletitle{TAGNN: Target attentive graph neural networks for session-based recommendation}. In \bibinfo{booktitle}{\emph{Proceedings of the 43rd International ACM SIGIR Conference on Research and Development in Information Retrieval}}. \bibinfo{pages}{1921–1924}.
\newblock


\bibitem[Yue et~al\mbox{.}(2023)]%
        {llamarec}
\bibfield{author}{\bibinfo{person}{Zhenrui Yue}, \bibinfo{person}{Sara Rabhi}, \bibinfo{person}{Gabriel de Souza Pereira~Moreira}, \bibinfo{person}{Dong Wang}, {and} \bibinfo{person}{Even Oldridge}.} \bibinfo{year}{2023}\natexlab{}.
\newblock \bibinfo{title}{LlamaRec: Two-Stage Recommendation using Large Language Models for Ranking}.
\newblock
\newblock
\showeprint[arxiv]{2311.02089}~[cs.IR]


\bibitem[Zhang et~al\mbox{.}(2023a)]%
        {llmfair}
\bibfield{author}{\bibinfo{person}{Jizhi Zhang}, \bibinfo{person}{Keqin Bao}, \bibinfo{person}{Yang Zhang}, \bibinfo{person}{Wenjie Wang}, \bibinfo{person}{Fuli Feng}, {and} \bibinfo{person}{Xiangnan He}.} \bibinfo{year}{2023}\natexlab{a}.
\newblock \showarticletitle{Is ChatGPT fair for recommendation? Evaluating fairness in large language model recommendation}. In \bibinfo{booktitle}{\emph{Proceedings of the 17th ACM Conference on Recommender Systems (RecSys)}}. \bibinfo{pages}{993–999}.
\newblock


\bibitem[Zhang et~al\mbox{.}(2023b)]%
        {attenmixer}
\bibfield{author}{\bibinfo{person}{Peiyan Zhang}, \bibinfo{person}{Jiayan Guo}, \bibinfo{person}{Chaozhuo Li}, \bibinfo{person}{Yueqi Xie}, \bibinfo{person}{Jae~Boum Kim}, \bibinfo{person}{Yan Zhang}, \bibinfo{person}{Xing Xie}, \bibinfo{person}{Haohan Wang}, {and} \bibinfo{person}{Sunghun Kim}.} \bibinfo{year}{2023}\natexlab{b}.
\newblock \showarticletitle{Efficiently leveraging multi-Level user intent for session-based recommendation via atten-mixer network}. In \bibinfo{booktitle}{\emph{Proceedings of the Sixteenth ACM International Conference on Web Search and Data Mining (WSDM)}}. \bibinfo{pages}{168–176}.
\newblock


\bibitem[Zhang et~al\mbox{.}(2023c)]%
        {mmsbr}
\bibfield{author}{\bibinfo{person}{Xiaokun Zhang}, \bibinfo{person}{Bo Xu}, \bibinfo{person}{Fenglong Ma}, \bibinfo{person}{Chenliang Li}, \bibinfo{person}{Liang Yang}, {and} \bibinfo{person}{Hongfei Lin}.} \bibinfo{year}{2023}\natexlab{c}.
\newblock \showarticletitle{Beyond Co-occurrence: Multi-modal Session-based Recommendation}.
\newblock \bibinfo{journal}{\emph{IEEE Transactions on Knowledge and Data Engineering}} (\bibinfo{year}{2023}), \bibinfo{pages}{1--12}.
\newblock
\urldef\tempurl%
\url{https://doi.org/10.1109/TKDE.2023.3309995}
\showDOI{\tempurl}


\bibitem[Zhang et~al\mbox{.}(2022)]%
        {pricerec}
\bibfield{author}{\bibinfo{person}{Xiaokun Zhang}, \bibinfo{person}{Bo Xu}, \bibinfo{person}{Liang Yang}, \bibinfo{person}{Chenliang Li}, \bibinfo{person}{Fenglong Ma}, \bibinfo{person}{Haifeng Liu}, {and} \bibinfo{person}{Hongfei Lin}.} \bibinfo{year}{2022}\natexlab{}.
\newblock \showarticletitle{Price DOES Matter! Modeling Price and Interest Preferences in Session-Based Recommendation}. In \bibinfo{booktitle}{\emph{Proceedings of the 45th International ACM SIGIR Conference on Research and Development in Information Retrieval}} \emph{(\bibinfo{series}{SIGIR '22})}. \bibinfo{pages}{1684–1693}.
\newblock


\bibitem[Zhang et~al\mbox{.}(2021)]%
        {LMasRS}
\bibfield{author}{\bibinfo{person}{Yuhui Zhang}, \bibinfo{person}{Hao Ding}, \bibinfo{person}{Zeren Shui}, \bibinfo{person}{Yifei Ma}, \bibinfo{person}{James Zou}, \bibinfo{person}{Anoop Deoras}, {and} \bibinfo{person}{Hao Wang}.} \bibinfo{year}{2021}\natexlab{}.
\newblock \showarticletitle{Language models as recommender systems: Evaluations and limitations}. In \bibinfo{booktitle}{\emph{NeurIPS 2021 Workshop on I (Still) Can't Believe It's Not Better}}.
\newblock


\end{thebibliography}
\end{document}